\documentclass{article}

\usepackage{PRIMEarxiv}

\usepackage[utf8]{inputenc} 
\usepackage[T1]{fontenc}    
\usepackage{hyperref}       
\usepackage{url}            
\usepackage{amsfonts}       
\usepackage{nicefrac}       
\usepackage{microtype}      
\usepackage{lipsum}
\usepackage{fancyhdr}       
\usepackage{graphicx}       
\graphicspath{{media/}}     
\usepackage{amsmath}
\usepackage{siunitx}
\usepackage{booktabs}
\usepackage{bm}
\usepackage{physics}
\usepackage{threeparttable}
\usepackage{gensymb}
\usepackage{subcaption}
\usepackage{balance}
\usepackage{cite}
\title{Bio-inspired Intelligence with Applications to Robotics: A Survey}
		
\author{
  Junfei Li, Zhe Xu, Danjie Zhu, Kevin Dong, Tao Yan, Zhu Zeng, Simon X. Yang \\
  School of Engineering, University of Guelph, Canada\\
  \texttt{\{jli64; zxu02; danjie; kdong; tyan03; zzeng05; syang\}@uoguelph.ca} \\
}

\begin{document}
\maketitle

\begin{abstract}
In the past decades, considerable attention has been paid to bio-inspired intelligence and its applications to robotics. This paper provides a comprehensive survey of bio-inspired intelligence, with a focus on neurodynamics approaches, to various robotic applications, particularly to path planning and control of autonomous robotic systems. Firstly, the bio-inspired shunting model and its variants (additive model and gated dipole model) are introduced, and their main characteristics are given in detail. Then, two main neurodynamics applications to real-time path planning and control of various robotic systems are reviewed. A bio-inspired neural network framework, in which neurons are characterized by the neurodynamics models, is discussed for mobile robots, cleaning robots, and underwater robots. The bio-inspired neural network has been widely used in real-time collision-free navigation and cooperation without any learning procedures, global cost functions, and prior knowledge of the dynamic environment. In addition, bio-inspired backstepping controllers for various robotic systems, which are able to eliminate the speed jump when a large initial tracking error occurs, are further discussed. Finally, the current challenges and future research directions are discussed in this paper.
\end{abstract}


\section{Introduction}
From the first stirrings of life, nature has been providing a suitable breeding ground for the intelligence of organisms. Biological intelligence enables organisms to adapt the extreme or changing environments. For instance,
a group of birds and fishes can efficiently sense the surrounding dynamic environments and take effective actions based on those inputs often with very simple mechanisms and with limited availability of information. Some species exhibit collective behaviors and can cooperatively accomplish tasks that are beyond the capabilities of a single individual under limited implicit communication.
Organisms with such beneficial traits can pass on these traits to offspring, exhibiting high adaptability to environments. 
The nervous system in the brain gives human abilities of feeling, thinking, and learning abilities.

Recently, there has been a general movement towards service-oriented robots that require the ability to adapt to complex dynamic situations and to handle various uncertainties. Due to the desirable properties of biological organisms, such as adaptability, robustness, versatility, and agility, the researchers have been trying to infuse robots with biological intelligence that will enable safe navigation and efficient cooperation among the autonomous robots in changing environments \cite{bekey2005autonomous}. The approaches inspired by biological intelligence are known as biologically inspired intelligence, which has been explored and studied for many years in robotics research \cite{Li_2019biosurvey}.  
The fundamental idea of biologically inspired intelligence is to incorporate useful biological strategies, mechanisms, and structures into the development of new methodologies and technologies to solve existing problems in a more efficient way than existing methodologies and technologies.
For instance, swarm intelligence and collective behaviors of living organisms have inspired the design of many robotics algorithms based on their biological strategies \cite{Pradhan_2020PSO,Huang_2013ACO}. The process of natural selection has inspired many computational models to optimize robot performances, such as genetic algorithm \cite{Roberge_2018GE,Hu2002ge} and differential evolution \cite{Huan_2018de}. The neural network algorithm, derived from neural science, has gained rising popularity among researchers around the world \cite{Guo_2019NN,Zhang_2019NN}. Biologically inspired intelligence algorithms were also integrated with various conventional algorithms to develop more efficient algorithms. For example, a knowledge based
genetic algorithm, which incorporated the domain knowledge into its specialized operators, was proposed to efficiently generate collision-free path of robots \cite{hu2004knowledge}. A neural network was used to convert the improved central pattern generator output to the foot trajectories of quadruped robots \cite{Zeng_2018}.
However, most bio-inspired studies are limited to conceptual or laboratory investigations or do not have much biological inspiration. Thus, the development of new intelligent strategies, algorithms and technologies are still highly needed, such as real-time collision-free navigation algorithms of individual robots or communication, coordination, and cooperation algorithms for multiple robotic systems, to accomplish multi-objective tasks in dynamic environments.

Bio-inspired neurodynamics models have been studied for real-time path planning and control of various robotic systems during the past decades \cite{Li_2019biosurvey}. The shunting neurodynamics model was derived from Hodgkin and Huxley’s membrane models for dynamic ion exchanges \cite{Grossberg_1973}. Based on the shunting neurodynamics model and its model variants, several new algorithms have been successfully developed for real-time path planning and control of various autonomous robots \cite{Yang_2001trajectory,Yang_2012real}. The definition of real-time is in the sense that the robot path planner and controller respond immediately to the dynamic environment, including the robots, targets, obstacles, sensor noise and disturbances. Many other model variants have been also developed for robot path planning and control. The additive model is computationally simpler and can generate real-time collision-free paths under most conditions \cite{Yang_2001trajectory,Yang_2003car}. The gated dipole model shows excellent performance in multi-robotic path planning and tracking control \cite{zhu_2003gate}.
Beyond the application of autonomous robots, bio-inspired neurodynamics models have been also widely applied to many other research fields, such as odor dispersion with electronic nose \cite{Pan_2009nose} and dynamic ginseng drying \cite{Martynenko_2006gensing}. These researches on agriculture have also been extended to biomedical and other industrial applications.

This paper focuses a comprehensive survey of the state-of-the-art research on bio-inspired neurodynamics models with their applications to path planning and control of autonomous robots. A detailed introduction of the shunting model and its variants are provided in this paper. Two main applications to robotic systems based on bio-inspired neurodynamics models are focused. The bio-inspired neural networks, in which each neuron is characterized by a neurodynamics model, is discussed for various robotic systems. The bio-inspired backstepping controllers that resolve the speed jump problem in tracking control is further discussed. The bio-inspired controllers have been successfully employed in tracking control and formation control. The pros and cons of different neurodynamics-based algorithms are also discussed in this paper.  The overall studies give an insight into neurodynamics models on autonomous robot applications, which could inspire potential ideas for future developments of novel intelligent bio-inspired path planning and control for diversified autonomous robotic systems. 

This paper is organized as follows: Section 2 introduces the background of bio-inspired neurodynamics models. Section 3 gives a survey on the path planning of various robots based on bio-inspired neurodynamics models. The applications of bio-inspired neurodynamics models to tracking control and formation control are presented in Section 4. Section 5 discusses the current challenges and future works. Some concluding remarks are finally summarized in Section 6.

\section{Bio-inspired Neurodynamics Models}
\label{sec:headings}
In this section, the originality of the shunting model is briefly described. Then, two model variants, the additive model and gated dipole model are also introduced.

\subsection{Originality}
\begin{figure}[htb]
  \centering
  \includegraphics[width=0.45\textwidth]{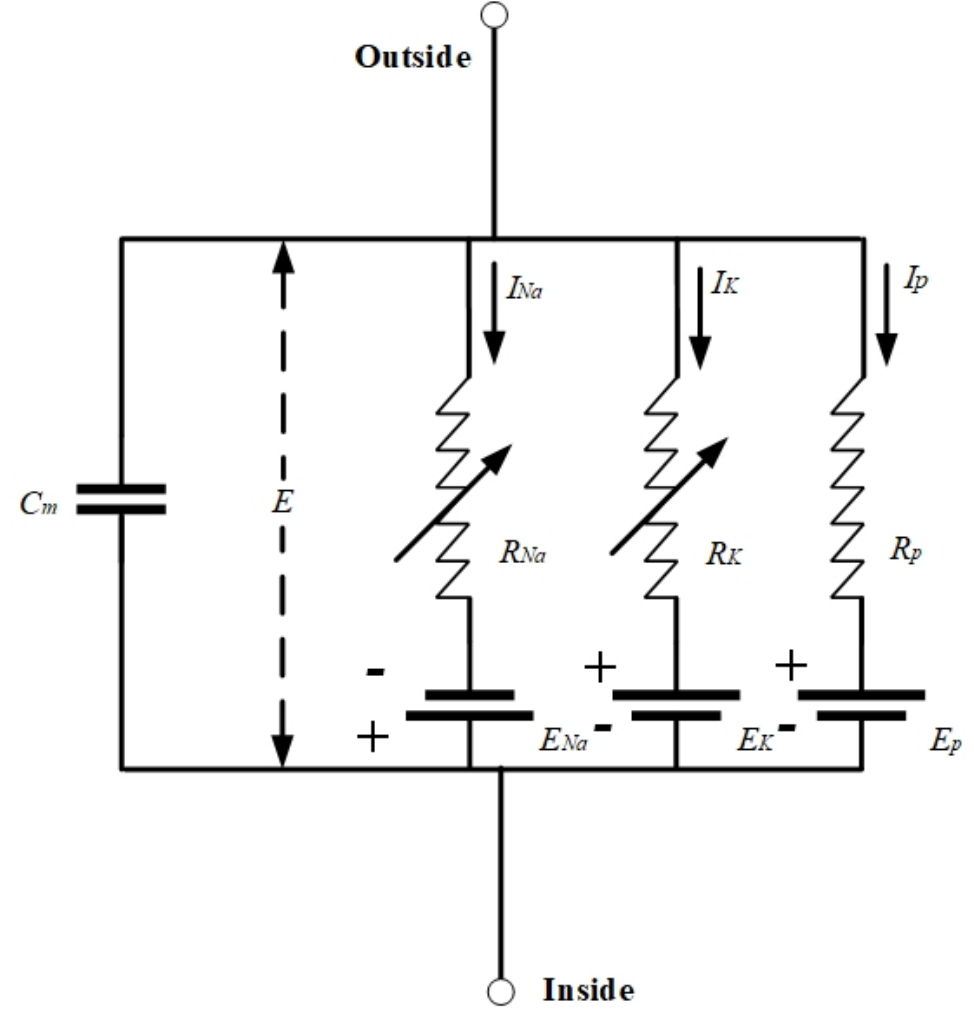}
  \caption{Electrical circuit representing membrane.}
  \label{fig:neuron_circuit}
\end{figure}

In 1952, an electrical circuit model was proposed by Hodgkin and Huxley to describe the action potential process in the membrane of neurons, based on experimental findings \cite{hodgkin1952quantitative}. The electrical behavior of the membrane can be represented by the circuit shown in \autoref{fig:neuron_circuit}. The dynamics of voltage across the membrane, $V_m$, is described using the state equation technique as
\begin{equation}
\begin{aligned}
C_{m} \frac{d V_{m}}{d t}=-\left(E_{p}+V_{m}\right) g_{p}+\left(E_{N a}-V_{m}\right) g_{N a}-\left(E_{K}+V_{m}\right) g_{K}
\label{eq:diffeq}
\end{aligned}
\end{equation}
where $C_{m}$ is the membrane capacitance; $E_{K}$, $E_{N a}$, and $E_{p}$ are the Nernst potentials (saturation potentials) for potassium ions, sodium ions, and passive leak current in the membrane, respectively; and $g_{K}$, $g_{N a}$ and $g_{p}$ represent the conductances of the potassium, sodium, and passive channels, respectively. Inspired from this membrane model for dynamic ion exchanges, Grossberg proposed a shunting model \cite{Grossberg_1973,Cohen_1983,Grossberg_1988}. By setting $C_{m} =1$ and substituting $u_{k}= E_p +V_m$, $A=g_p$, $B=E_{Na}+E_p$, $D=E_k-E_P$, $S^e_k=g_{Na}$, and $S^i_k=g_{K}$ in \autoref{eq:diffeq}, a shunting equation is obtained as \cite{_men_1990,_Omen_1990b} 
\begin{equation}
\frac{d x_{k}}{d t}=-A x_{k}+\left(B-x_{k}\right) S_{k}^{e}-\left(D+x_{k}\right) S_{k}^{i}
\label{eq:shunnting}
\end{equation}
where $x_{k}$ is the neural activity (membrane potential) of the $k$-th neuron; $A$, $B$, and $D$ are nonnegative constants representing the passive decay rate, the upper and lower bounds of the neural activity, respectively; $S^e_k$ and $S^i_k$ are the excitatory and inhibitory inputs to the neuron, respectively. 
In the shunting model, $B$ and $D$ are not essential factors because the neural activity is the relative values between the boundary lines. Only parameter $A$ determines the model dynamics. However, $A$ can be chosen in a very wide range. Thus, the shunting model is not very sensitive to the model parameters \cite{Yang_2001trajectory}.

\autoref{eq:shunnting} shows that the increase of activity $x_k$ depends on the positive term $(B-x_k)S^e_k$ that relies on both the excitatory input $S^e_k$ and the difference of neural activity to its upper bound $(B-x_k)$. Therefore, the increases of $x_k$ become slower as the value of $x_k$ is closing to the upper bound $B$. If the value of $x_k$ equals to $B$, the $(B-x_k)$ term becomes zero, and positive term has no effect no matter how big the excitatory input $S^e_k$ is. In the case that the value of $x_k$ is greater than $B$, the $(B-x_k)$ term becomes negative, then the positive term becomes negative, the excitatory input will decrease the activity $x_k$ until it is not higher than $B$. Therefore, $B$ is the upper bound of the neural activity $x_k$. The same for the negative term $(D+x_{k})S^i_k$, which guarantees that the neural activity $x_k$ is always greater than the lower bound $-D$. Thus, the neural activity $x_k$ is bounded between the $[-D, B]$ region under various inputs conditions. The shunting model has been studied to understand the adaptive behaviors of individuals in dynamic and complex environments \cite{Grossberg_1973}. Many achievements have been accomplished in the past decades, such as, machine vision, sensory motor control, and many other areas \cite{_men_1990,Grossberg_1988}. In the field of robotics, the shunting model has been wildly used in path planning, tracking control, hunting, cooperation of various autonomous robots \cite{Yang_2001trajectory,Yang_2002tracking,Ni_2017,Ni_2015formation}. 
\subsection{Model Variants}

If the excitatory and inhibitory inputs in \autoref{eq:shunnting} are lumped together and the auto-gain control terms are removed, then \autoref{eq:shunnting} can be written into a simpler form
\begin{equation}
\frac{d x_{k}}{d t}=-Ax_{k}+S_k
\label{eq:addS}
\end{equation}
where $S_k$ is the total inputs of the $k$-th neuron. Then, \autoref{eq:addS} is rewritten as:
\begin{equation}
\frac{d x_{k}}{d t}=-Ax_{k}+I_{k}+\sum_{l=1}^{N}w_{kl}f(x_l)
\label{eq:additive}
\end{equation}
where $w_{kl}$ is the connection weight from the $l$-th neuron to the $k$-th neuron; $f()$ is an activation function; $I_k$ represents the external input to the $k$-th neuron; and $N$ is the total number of neurons in the neural network. In most situations, the additive model is computationally simpler and can also generate the real-time collision-free path for robots. However, the shunting model has two important advantages. Firstly, the shunting model in \autoref{eq:shunnting} has excitatory and inhibitory auto-gain control terms, $(B-x_k)$ and $(D+x_k)$, respectively, which give the shunting model the dynamic responsive ability to input signals. The shunting model is more sensitive to the changes of inputs \cite{Yang_2001trajectory}. Nevertheless, the dynamics of the additive model may saturate in some situations. Secondly, the shunting model is bounded between the upper bound $B$ and lower bound $-D$, whereas the additive model is bounded only by limiting the input signals. The additive models have been widely applied to artificial vision, learning-based algorithms, and other research fields \cite{Grossberg_1988}. Owning to the simple computation process, even the limitations of the additive model exist, the additive model has been also applied to replace the shunting model in many situations \cite{Yang_2001trajectory,Yang_2003car}.
\begin{figure}[htb]
  \centering
  \includegraphics[width=0.32\textwidth]{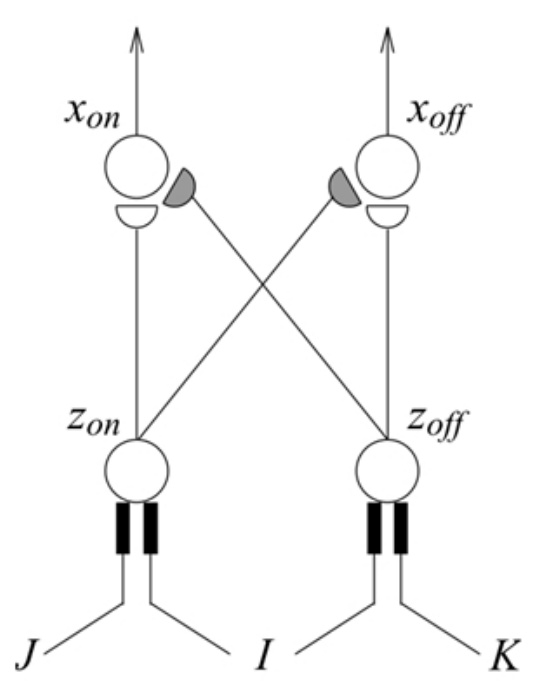}\hspace{1mm}
  \caption{A gated dipole model.}
  \label{fig:gated}
\end{figure}

Another essential neurodynamics model is the gated dipole model, which is shown in \autoref{fig:gated}. A basic gated dipole model is consisted of the opponent on-channel and off-channel. An arousal signal $I$ can stimulate both on- and off- channels. The extra inputs $J$ and $K$ stimulate the on-channel and off-channel, respectively. The dynamics of the available transmitters are characterized by
\begin{equation}
\begin{array}{l}
\frac{dz_\textit{on}}{dt}=\alpha\left(\beta-z_\textit{on}\right)-\gamma(I+J)z_\textit{on}
\end{array}
 \label{eq:gated channelON}
\end{equation}
\begin{equation}
\begin{array}{l}
\frac{d z_\textit{off}}{dt}=\alpha\left(\beta-z_\textit{off}\right)-\gamma(I+K)z_\textit{off}
\end{array}
 \label{eq:gated channelOFF}
\end{equation}
where $z_\textit{on}$ and $z_\textit{off}$ are the number of available transmitters in the on- and off-channels, respectively; $\alpha$ and $\gamma$ are the transmitter production and depletion rates, respectively; and $\beta$ represents the total amount of transmitter. The on-cells receive excitatory inputs from the on-channel, while receive inhibitory inputs from its opponent channel (off-channel). Similar to the off-channel, the off-cells receive excitatory inputs from the off-channel, while receive inhibitory inputs from its opponent channel (on-channel). Thus, the dynamics of the on- and off-channels are characterized by the following shunting equations
\begin{equation}
\begin{aligned}
\frac{d x_\textit{on}}{dt}=&-Ax_\textit{on}+\left(B-x_\textit{on}\right)(I+J)z_\textit{on}-\left(D+x_\textit{on}\right)(I+K)z_\textit{off}
\end{aligned}
\end{equation}
\begin{equation}
\begin{aligned}
\frac{dx_\textit{off}}{dt}=&-Ax_\textit{off}+\left(B-x_\textit{off}\right)(I+K) z_\textit{off}-\left(D+x_\textit{off}\right)(I+J)z_\textit{on}
\end{aligned}
\end{equation}
where $x_\textit{on}$ and $x_\textit{off}$ are the activities of the on-channel and the off-channels, respectively. In the on-channel, the available transmitters decrease exponentially to a plateau when the extra light $J$ is on, and goes back to its initial resting level in the same manner after the offset of the light. The available transmitter in the off-channel stays constant since there is no change of light. In the on-channel, when the extra light $J$ turns on, there is more available transmitter depleted, and the response of the on-cell initially overshoot. However, after the onset of light, the available transmitter decreases exponentially due to temporal adaptation, the activity of on-cell decays exponentially to a plateau. At the offset of the extra input, the on- and off-channels have the same input $I$, while the available transmitter in the on-channel is used up by the depletion during the on-light, the activity of the on-channel appears a rebound which is called antagonistic rebound. After the extra light-off, due to temporal adaption, the available transmitter rises exponentially to its resting level, the activity of the on-channel climbs exponentially back also. The gated dipole model was successfully used to explain many biological phenomena that involve agonist and antagonist interaction \cite{Oh_2017gated}, and had applications for robotic research of path planning and tracking control \cite{zhu_2003gate,Yang2004}.

\section{Path Planning}
\label{sec:others}
\begin{figure}[htb]
  \centering
  \includegraphics[width=0.40\textwidth]{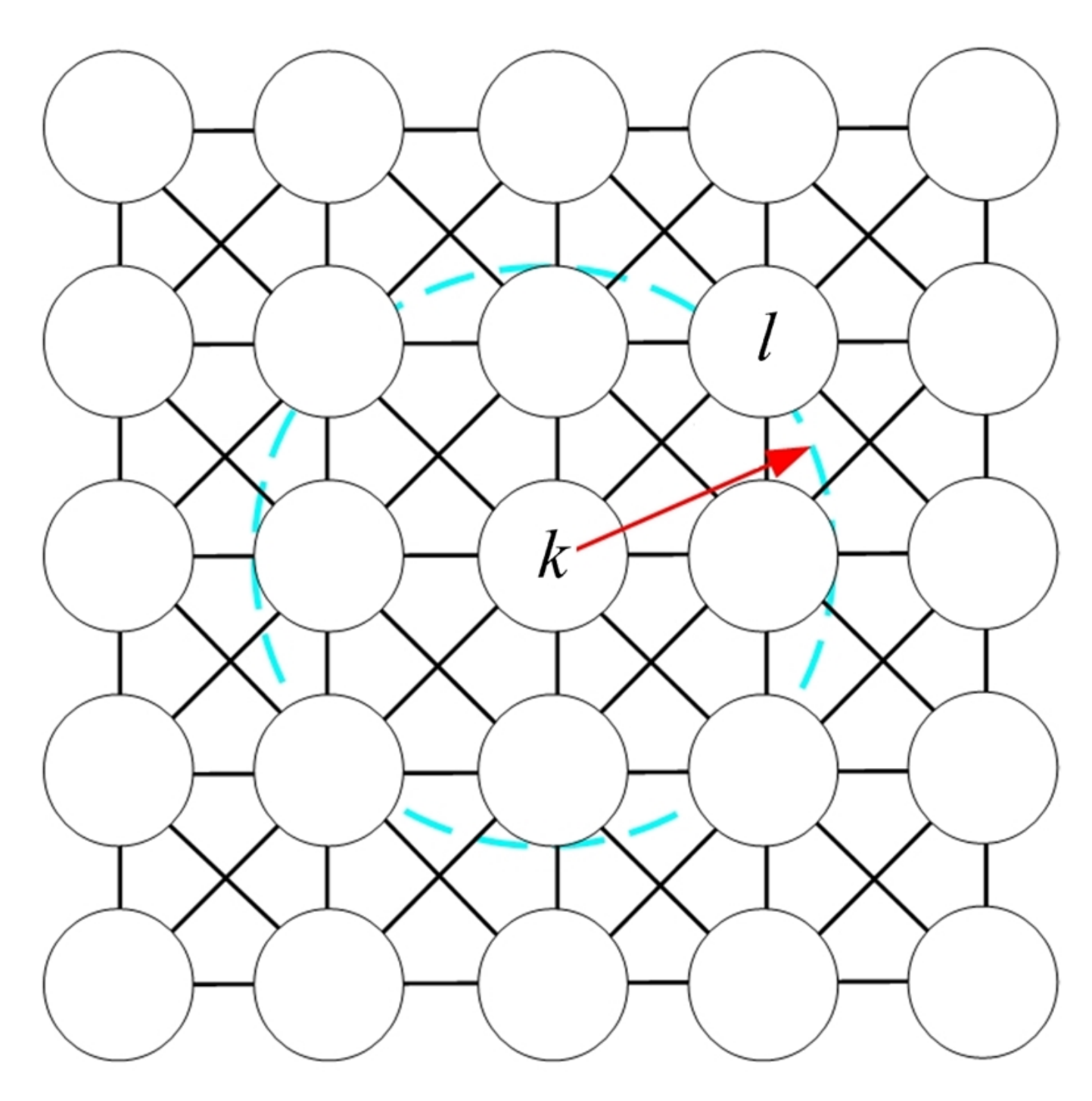}
  \caption{An example of the bio-inspired neural network.}
  \label{fig:BINN}
\end{figure}

A basic path planning problem can be defined as: given a work environment with obstacles, the target, and the initial robot position, a collision-free path should be generated from the initial position to the target. The bio-inspired neurodynamics model has been wildly used in real-time path planning without any learning procedures and any prior knowledge of the dynamic environment. The key point of the neurodynamics-based path planning approach is to represent the work environment as one-to-one corresponding to the neurons in the neural network. The dynamics of each neuron in the neural network are characterized by \autoref{eq:shunnting}. An example of a neural network is shown in \autoref{fig:BINN}. Thus, the neural activity for the $k$-th neuron is obtained by 
\begin{equation}
\begin{aligned}
\dv{x_k}{t}=-Ax_k+(B-x_k)\Bigg([I_k]^++\sum^n_{l=1}w_{kl}[x_l]^+\Bigg)-(D+x_k)[I_k]^- 
\end{aligned}
\end{equation}
where $x_l$ represents the activity value of those neighboring neurons; $n$ is the number of neighboring positions of the $k$-th neuron; $[a]^+$ is a linear-above-threshold function defined as $[a]^+=\max\left\{a,0\right\}$; and  $[a]^-$ is defined as $[a]^-=\max\left\{-a,0\right\}$. In the bio-inspired neural network, the excitatory input $S^e_k$ is consisted of two parts, $[I_k]^+$ and $\sum^n_{j=1}w_{kl}[x_l]^+$, where $[I_k]^+$ is the external input from targets, and $\sum^n_{j=1}w_{kl}[x_l]^+$ is the internal input through the propagation of the positive activity from its neighborhoods. The inhibitory input $S^i_k$ has only external input $[I_k]^-$, which is from the obstacle, and only has local effects (no negative activity propagation). Thus, the target has maximum and positive neural activity, which could globally propagate through the neural network to attract the robot, while the obstacles have only local effects without propagating.
The path selection rule of the individual robot can be defined as: the next move position of the robot is the maximum neural activity of its current neuron’s neighbors. After robots move to the next position, the next position becomes a new current position until the current position is the location of the target. The robot would never choose the position of an obstacle to be the next movement due to the negative neural activity of obstacles. Thus, the robot is able to avoid collisions and move to the target. It is important to note that the path planning process is without any learning procedures and any prior knowledge of the dynamic environment. Due to the real-time performance and computational efficiency, bio-inspired neural network path planning approaches have been developed for various robot systems. In this section, based on the different types of robots, three categories are divided: mobile robots, cleaning robots, and underwater robots.

\subsection{Mobile Robots}
Path planning of mobile robots has received a lot of interest because mobile robots have been participating in human life. In this section, two main challenges of mobile robot path planning are focused: real-time collision-free navigation and the cooperation of the multi-robot systems. In addition, many developed model variants are also discussed for robot path planning.

\subsubsection{Navigation}
\begin{figure}[htb]
  \centering
  \includegraphics[width=0.95\textwidth]{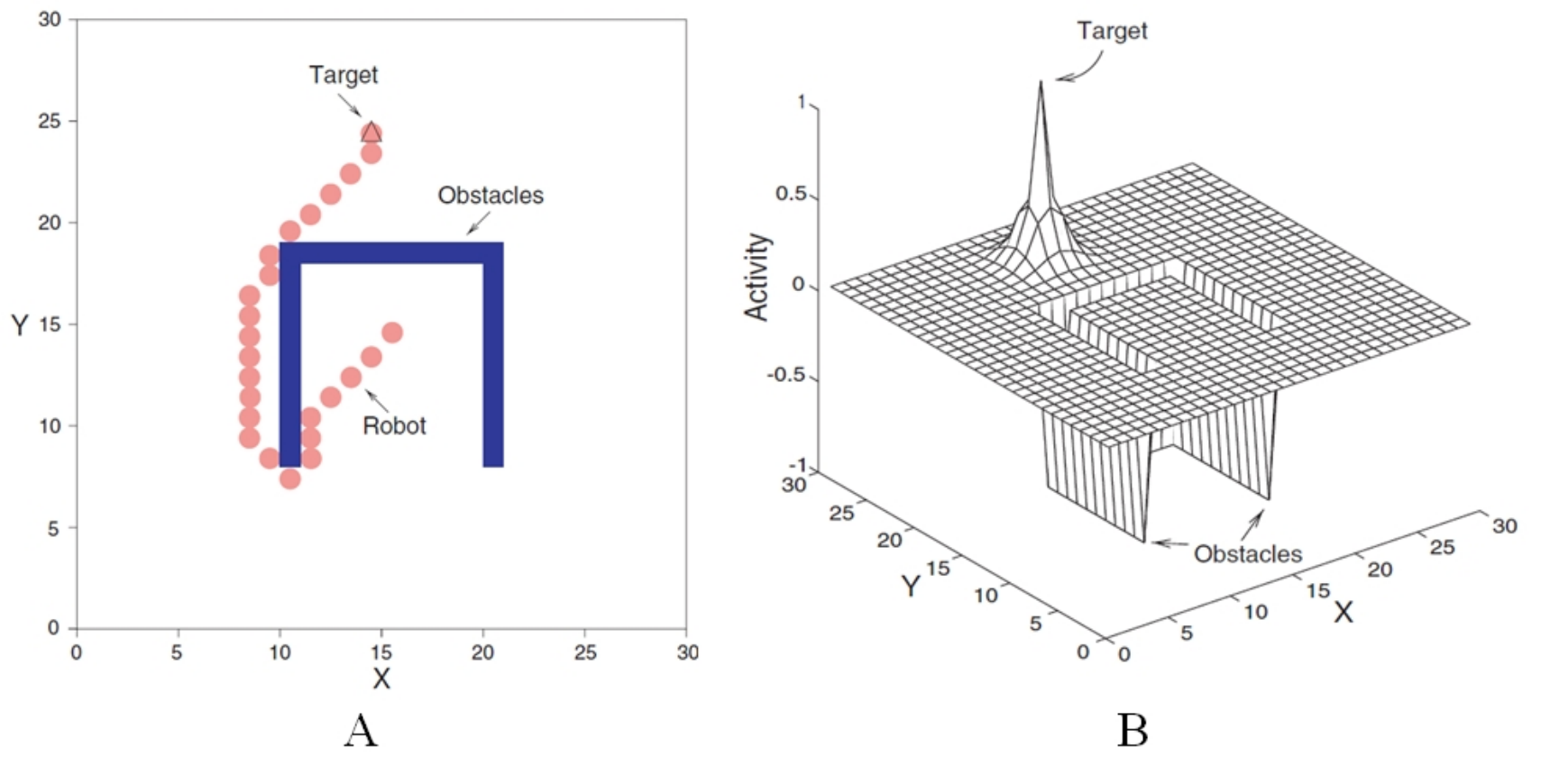}\hspace{1mm}%
  \caption{Path planning of a mobile robot to avoid local minima with concave obstacles. A: the robot path; B: the landscape of neural activity \cite{Yang_2001trajectory}.}
  \label{fig:motionplaning}
\end{figure}

The first bio-inspired neural network framework was proposed by Yang and Meng for the mobile robot path planning \cite{Yang_2000path}. Many remarkable achievements in mobile robot path planning have been achieved \cite{Yang20023,Yang_2001trajectory,Yang_20026}. Due to the global effects of positive neural activity from the target, the robot is not trapped in the undesired local minima. \autoref{fig:motionplaning} shows an example of path generation of a mobile robot to avoid local minima. The robot is not trapped in a set of concave obstacles and move to the target position.

\begin{figure}[htb]
  \centering
  \includegraphics[width=0.95\textwidth]{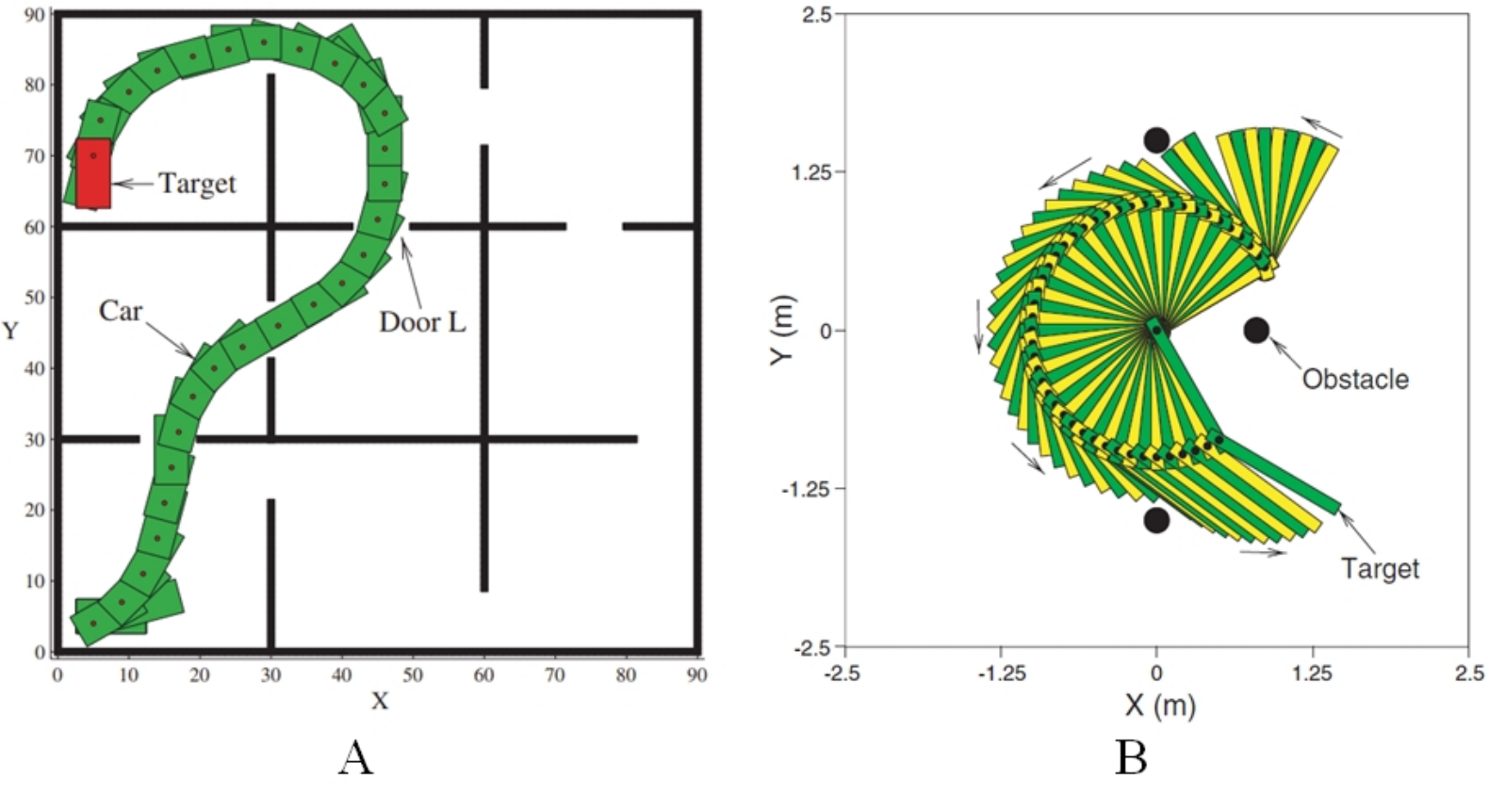}\hspace{1mm}%
  \caption{Examples of a nonholonomic car-like robot and a manipulator robot. A: robot motion when the door is opened \cite{Yang_2003car}; B: simple planar robot avoiding obstacles \cite{Yang_2001trajectory}.}
  \label{fig:car}
\end{figure}

Some researchers consider the different types of robots in the application to navigation. A nonholonomic car-like robot was studied by Yang \textit{et al.}\cite{Xianyi_Yang1999car,Yang2000car,Yang_2003car} for real-time collision-free path planning. The simulation results showed the car-like robots performed well in parallel parking, navigation in several deadlock situations, and sudden environmental changes conditions. 
In a house-like environment as shown in \autoref{fig:car}A, the robot moved to the target along the shortest path in case that the door is opened. When the door is closed, the robot travels a much longer path to reach the target without any learning procedures. The robot is capable of reaching the target along the shortest path without any collisions, without violating the kinematic constraint, and without being trapped in deadlock situations.

In addition, Yang and Meng developed the bio-inspired neural network for robot manipulators \cite{Yang_2001trajectory}. The joint space of the robot manipulators was corresponded to the bio-inspired neural network, in which neurons were characterized by the shunting model or the additive model. \autoref{fig:car}B shows the trajectory of robot manipulators avoiding obstacles. In addition, a virtual assembly system was proposed by Yuan and Yang for assisting product engineers to simulate the assembly-related manufacturing process \cite{Xiaobu_Yuan_2003Virtual}. 

An improved bio-inspired neural network based on scaling terrain was proposed by Luo \textit{et al.}\cite{Luo_2019scaple} for reducing the calculation complexity. This multi-scale method mentioned better performance in terms of time complexity. However, the simulation experiments do not give the criteria for choosing the parameter of coarse-scale and fine-scale maps. Ni \textit{et al.}\cite{Ni_2016Qlearning} used a bio-inspired neurodynamics model as the reward function for the Q-learning algorithm, which can reduce the eﬀect of the reward function on the convergence speed. 

Some researchers pointed out that if the planned path is too close to the obstacles, it is dangerous for robot navigation. A dynamic risk level was incorporated to the shunting neurodynamics model to reduce the probability of collision in the dynamic obstacle avoidance task \cite{Jianjun_Ni_2014safe}. In addition, a novel 3-D neural dynamic model was proposed and expected to obtain the safety-enhanced trajectory in the work space considering of minimum sweeping area \cite{Chen_2017safe}. A safety consideration path planning can be implemented by setting a constant value $\sigma$ to inhibitory inputs in \autoref{eq:shunnting}. The safety consideration shunting equation is obtained by \cite{Yang_1998safe,Yang_2000safe}
\begin{equation}
\begin{aligned}
\frac{\mathrm{d} x_{k}}{\mathrm{~d} t}=-A x_{k}+\left(B-x_{k}\right)\left(\left[I_{k}\right]^{+}+\sum_{l=1}^{n} w_{kl}\left[x_{l}\right]^{+}\right)-\left(D+x_{k}\right)\left(\left[I_{k}\right]^{-}+\sum_{l=1}^{n} v_{kl}\left[x_{l}-\sigma\right]^{-}\right)
\end{aligned}
\label{eq:safe}
\end{equation}
where parameter $\sigma$ is the threshold of the inhibitory lateral neural connections. In \autoref{eq:shunnting}, the inhibitory input $S^i_k$ is only from the obstacles. However, in the safety consideration model, the inhibitory input $S^i_k$ is consisted of two parts: $\left[I_{k}\right]^{-}$ and $\sum_{l=1}^{n} v_{kl}\left[x_{l}-\sigma\right]^{-}$.The $\sum_{l=1}^{n} v_{kl}\left[x_{l}-\sigma\right]^{-}$ term guarantees that the negative activity 
propagates to a small region due to the threshold $\sigma$ of the inhibitory lateral neural connections. Thus, there is a small negative neural activity region surrounding the obstacles, and the robot is able to keep a safe distance from obstacles to avoid possible collisions.

Many variants of the bio-inspired neurodynamics models have been developed to deal with different situations.
The additive model generates the real-time collision-free robot paths under most conditions \cite{Yang_2001trajectory}. Even the computation of the additive model is simpler, the real-time performance of the additive model could be saturated in many situations. 
A similar neural network model was proposed by Glasius \textit{et al.}\cite{Glasius_1995GBNN} for real-time trajectory generation. Even Glasius's model had limitations with fast dynamic systems, Glasius bio-inspired neural network models have been used in underwater robots \cite{Sun_2019GBNN,Chen_2018GBNN,Chen_2019GBNN}. 
Inspired by the bio-inspired neural network model, a distance-propagating dynamic system was proposed that can efficiently propagate the distance instead of the neural activity from the target to the entire robot work space \cite{Willms_2006distance}. After that, Willms and Yang designed the safety margins around obstacles. The robots not only avoid obstacles but also keep a safe distance between the obstacles \cite{Willms_2008safe}.
Based on Willms and Yang's previous work, a shortest path neural networks model was proposed by Li \textit{et al.}\cite{Shuai_Li_2007spnn} for generating the globally shortest path.
A modified pulse-coupled neural network was proposed by Qu \textit{et al.}\cite{Hong_Qu_2009pulse,Hong_Qu_2013pulse} for real-time collision-free path planning. The computational complexity of the algorithm was only related to the length of the shortest path.
In addition, an improved Hopfield-type neural network model was proposed by Zhong \textit{et al.}\cite{Yongmin_Zhong_2008Hopfield} for easily responding the real-time changes in dynamic environments.
A padding mean neurodynamics model was proposed by Chen \textit{et al.}\cite{Chen_2020PMNDM} for the reasonable path generation in both static and dynamic varying environments.

\subsubsection{Cooperation of Multi-robotic Systems}
A team of robots would work together to accomplish an assigned task rapidly and efficiently. In many challenging applications such as search and rescue operations, security surveillance and safety monitoring, a multi-robotic system has obvious advantages than a single robotic system. The key challenge of multi-robotic systems in dynamic environments is to infuse these robots with biologically inspired intelligence that will enable efficient cooperation among the autonomous robots, and successful completion of designated tasks in changing environments. 

For the multiple targets path planning, an online solution based on the bio-inspired neural network was proposed by Bueckert \textit{et al.}\cite{Bueckert_2007path} in static and dynamic environments. However, the task assignment approach was very simple, as the robot visited the target, this target was removed from the visit list. Thus, the robot was hard to find the optimal visit sequence of targets. A novel hybrid agent framework was proposed by Li \textit{et al.}\cite{Howard_Li_2008path} for real-time path planning to multi-robotic systems considering many moving obstacles. In this work, an improved shunting equation was proposed by setting safety margins for the robots and the moving obstacles. The robots are able to predict the movement of obstacles and avoid any collision. Nanoassembly planning creates enormous potential in a vast range of new applications. An integrated method based on the shunting model was proposed to generate collision-free paths of multi-robotic nanoassembly \cite{Yuan_2007nano}. The tasks of the multi-robotic nanoassembly planning were a continuous process considering the environmental uncertainty.

\begin{figure}[htb]
  \centering
  \includegraphics[width=0.95\textwidth]{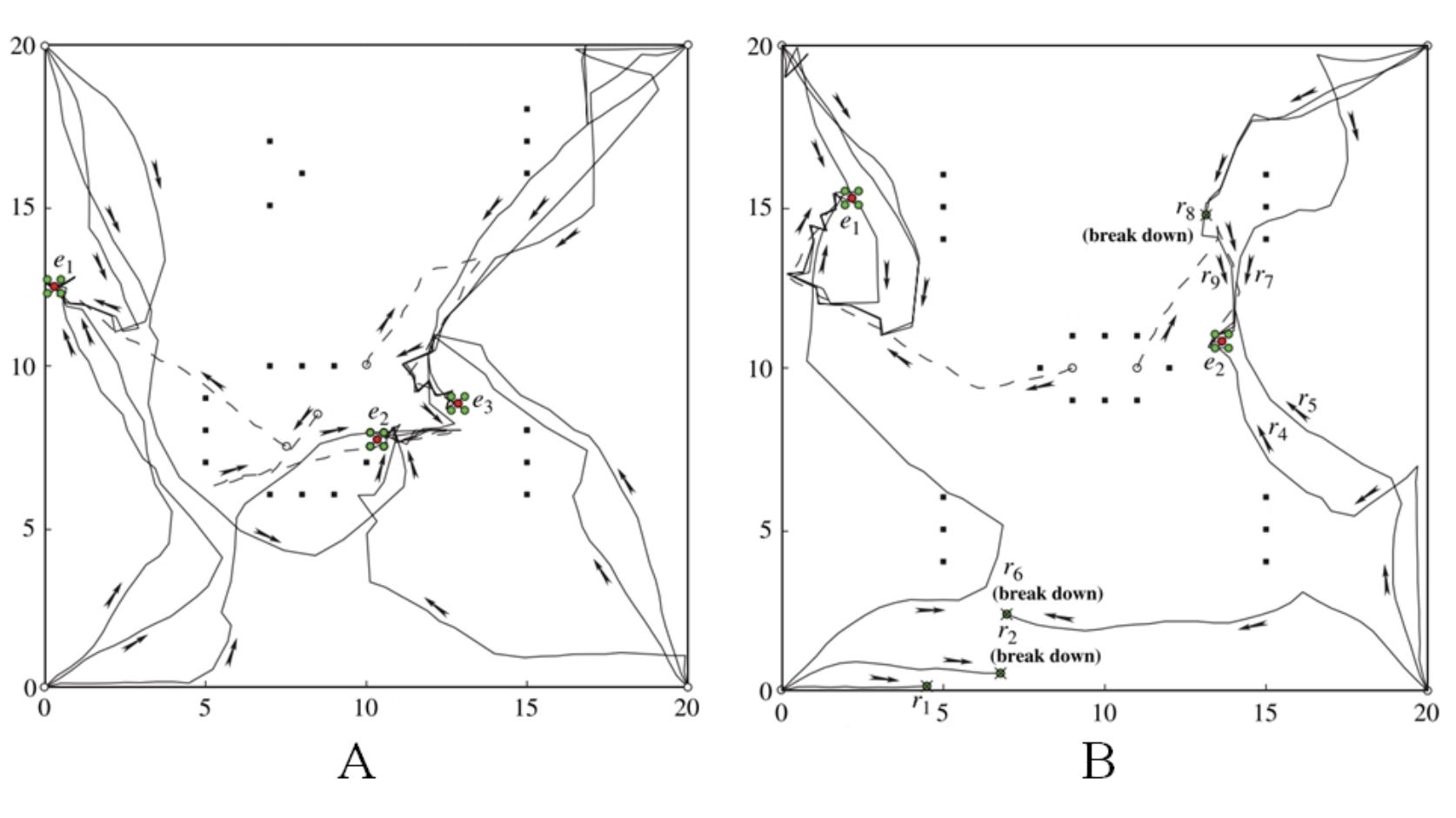}\hspace{1mm}%
  \caption{Examples of hunting tasks. A: multiple evaders need to be hunted; B: some robots break down \cite{Jianjun_Ni_2011hunting}.}
  \label{fig:hunting}
\end{figure}

If the robotic systems need to track the moving targets, an important influence of the algorithms is the relative moving speed between the target and robot \cite{Anmin_Zhuanlyz}. If the speed of robotic systems is much lower than the target, the robotic systems need to corporately track the target, otherwise the robot will never catch the target. A real-time cooperative hunting algorithm was proposed by Ni and Yang base on shunting neurodynamics models \cite{Jianjun_Ni_2011hunting}. In this hunting task study, the robots had no previous knowledge about the environment and locations of evaders.
It is important to note that the difference between tracking a moving target and the hunting algorithm is that the evader in hunting problems has some intelligence to escape from the hunt of pursuer robots. \autoref{fig:hunting}A  shows the hunting process considering many evader robots. Compared with other hunting algorithms, the hunting algorithm based on bio-inspired neurodynamics still works efficiently when some hunting robots are broken. \autoref{fig:hunting}B shows the hunting process that some robots are broken.

\subsection{Cleaning Robots}

The cleaning tasks require the robot to pass through every area in the work environment. The task requirement is the same as the complete coverage path planning (CCPP), which is a special type of path planning in 2-D environments. The CCPP can be also applied to many other robotic applications, such as painter robots, demining robots, lawnmowers, automated harvesters, agricultural crop harvesting equipment, windows cleaners, and autonomous underwater covering vehicles \cite{Yang_2004ccpp,Godio_2021ccppuav}. In the bio-inspired neural network, the unclean areas are set as targets, which globally attract the robot. The obstacles have only local effects, which avoids robot collisions \cite{Luo_2002ccpp,Yang_2002ccpp,Luo_2002sweeping}. As the cleaning robot works, the unclean areas become clean and the excitatory input of the clean area becomes zero. Thus, the landscape of neural activity dynamically changes with the change of the unclean areas, obstacles, and other robot position. For any current position of the robot, the next robot position $p_n$ is obtained by
\begin{equation}
p_{n} \Leftarrow x_{p_{n}}=\max \left\{x_{l}+c y_{l}, l=1,2, \cdots, n\right\}
\label{CCPP}
\end{equation}
where $c$ is a positive constant and $y_l$ is a monotonically increasing function of the difference between the current to next robot moving directions. Compared with path planning and CCPP problems, the main difference is that many target positions might attract the cleaning robot because all unclean areas are set as targets. Thus, the turning numbers of clean robots might increase significantly. Function $y_l$ is designed to reduce the turning numbers. If the robot goes straight, $y_l$ =1; if goes backward, $y_l$ = 0. Thus, the cleaning robot tends to go straight.

In this section, based on the previous knowledge of the environment, the research fields of cleaning robots using the neurodynamics model are categorized as: completed known environment and unknown environment.

\subsubsection{Completed Known Environment}
\begin{figure}[htb]
  \centering
  \includegraphics[width=0.95\textwidth]{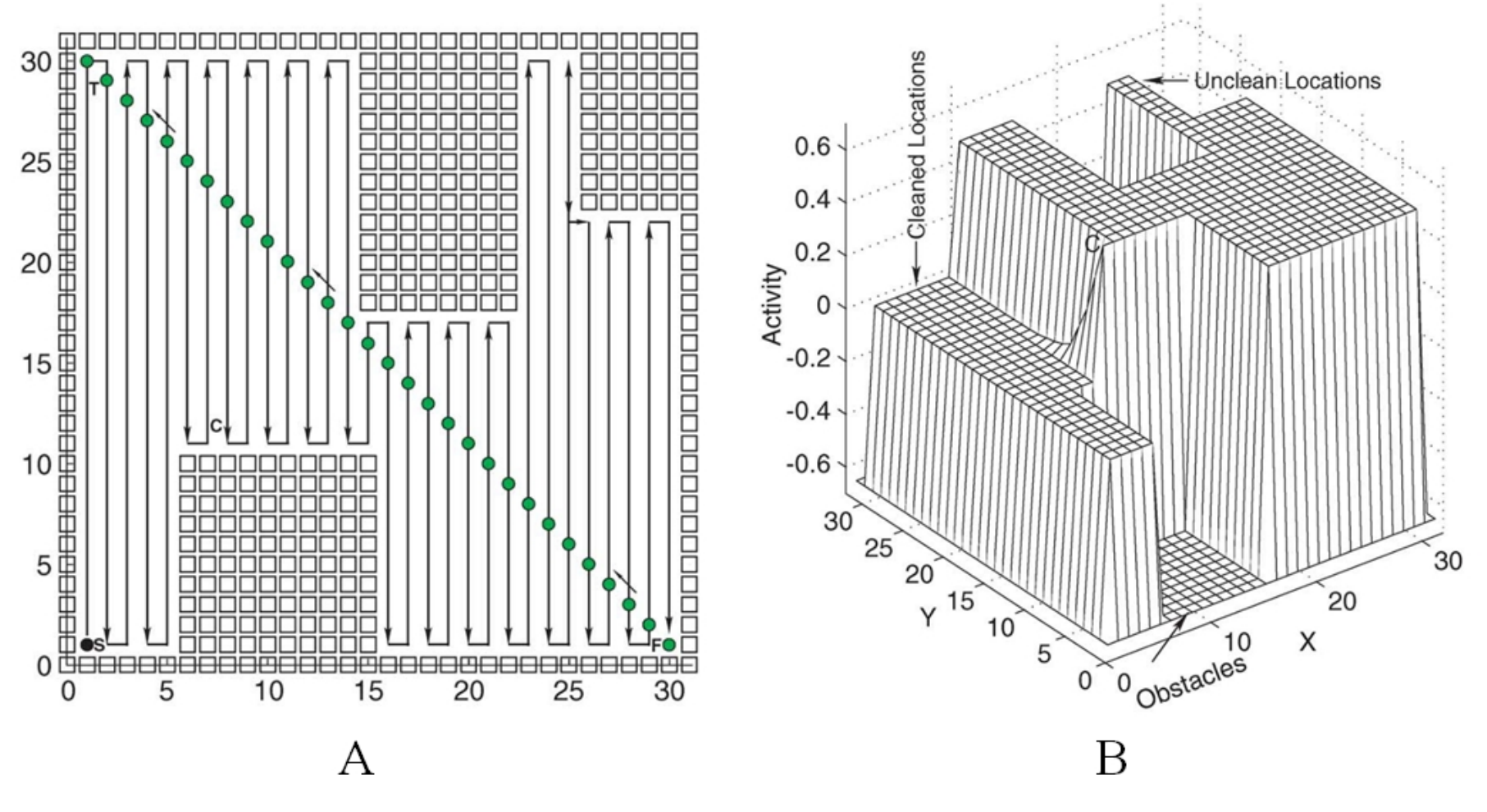}\hspace{1mm}%
  \caption{CCPP in a completely known environment. A: the generated robot path; B: the neural activity landscape when the robot reaches point C \cite{Luo_2008ccppUN}.}
  \label{fig:ccpp}
\end{figure}

\autoref{fig:ccpp} shows the neurodynamics-based CCPP in a completely known environment. The neurodynamics model can work efficiently in the dynamic environment, so even considering sudden change environment and moving obstacles in the environment, the cleaning robots can still work efficiently\cite{Luo_2002ccpp,Yang_2002ccpp,Yang_2004ccpp}. In order to improve the computational complexity, a discrete bio-inspired neural network was proposed to convert to the shunting equation a difference equation \cite{Zhang_2017ccpp}. 

One CCPP challenging problem is the deadlock situation. The deadlock area is a specific situation that the cleaning robot is trapped in a position where all of the neighborhood areas have been covered, but the work environment is still unclear. If the cleaning robot moves to deadlock areas, the cleaning robot is unable to escape from the deadlock areas without any interventions. A dynamic neural neighborhood analysis for deadlock avoidance was proposed based on the characteristics of deadlock areas \cite{Luo_2002ccpp}. The robot can recognize whether the current position is the deadlock point. If the current position is a deadlock point, the connection weights of the neural network were changed to generate a path to escape this deadlock point. 

\subsubsection{Unknown Environment}

In order to deal with CCPP in the unknown environment, the cleaning robots are typically required to build a surrounding map with a very limited time range \cite{Luo_2008ccppUN}. The onboard sensors have been widely used for robot navigation with a limited reading range. Thus, the key challenge of CCPP in unknown environments is to design the map-building algorithm and combine it with previous coverage algorithms studies. Combing with the sensor detection, an improved CCPP algorithm based on the neurodynamics model was developed in unknown environments \cite{Luo2005ccppUN,Luo_2006ccppUN}. The robots move to the nearest unclean areas and detect the environment until the cleaning task is finished. A real-robot platform iRobot Create 2 was used to test the proposed algorithm in unknown environments \cite{Luo_2017ccppUN}. The actual cleaning robots testing showed that the effectiveness of the proposed algorithm, in which the robotic systems could cooperatively work together in a large and complex environment.

\subsection{Underwater Robots}
 The autonomous underwater vehicle (AUV) or unmanned underwater vehicle (UUV) have been studied in a variety of tasks such as underwater rescue, data collection, and ocean exploration. In addition, some bionic robots are also studied, such as robotic fish \cite{Yu_2019fish,Yu_2021fish}. Unlike the work environment of mobile robots or cleaning robots, the underwater environment is more complex and uncertain. Firstly, based on the 2-D neural network structure, a 3-D grid-based neural network is typically required to represent the underwater environment. Secondly, the effect of the ocean or river currents is necessary to consider. Finally, the robots work in underwater environments, facing many uncertainties, such as some robots broken down. Based on different task requirements, three major research fields of underwater robots using the neurodynamics model are studied in this section.

\subsubsection{Navigation}
\begin{figure}[htb]
  \centering
  \includegraphics[width=0.55\textwidth]{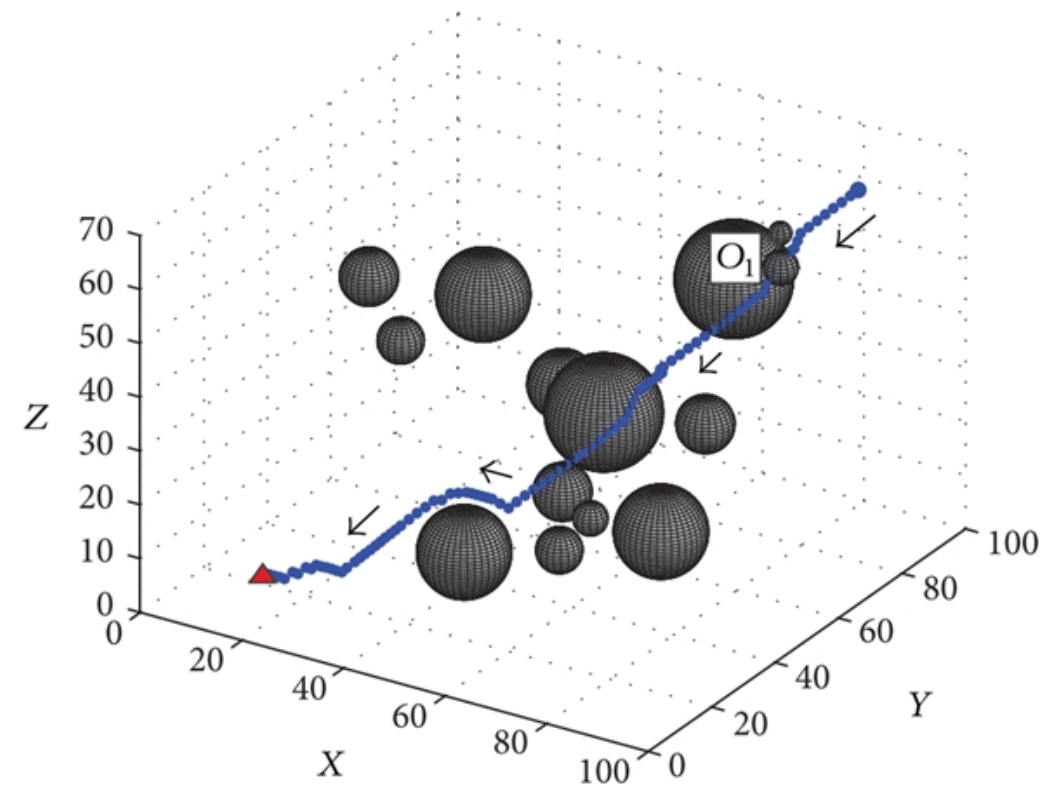}
  \caption{An example of path planning in 3-D underwater environments \cite{Ni_2017}.}
  \label{fig:3Dpathplanning}
\end{figure}

For the underwater environment, the neural network architecture needs to be extended to the 3-D environment, where more complex topography of randomly distributed obstacles is involved. 
\autoref{fig:3Dpathplanning} shows a typical AUV path planning in 3-D underwater environments. In 2-D neural network architecture, each neuron connects with 8 neighborhood neurons, whereas, in the 3-D neural network, each neuron connects with 26 neighborhood neurons \cite{Yan_2013}. Thus, the computation complexity dramatically increased. 
In order to improve the efficiency in the 3-D underwater environment, a dynamic bio-inspired neural network was proposed to guide the movement of AUV in large unknown underwater environments \cite{Ni_2017}. A virtual target selection approach was applied to search the path and avoid dead loop situations. Since the large unknown environment is partitioned into small portions centered with moving AUV, and the bio-inspired neural network only deals with this small range, the path can be calculated relatively fast. However, the dynamic bio-inspired neural network misses the best route in certain circumstances, which could waste the power of the AUV.

The environmental disturbances of the underwater area, such as currents, create inevitable influences on the AUV path planning. A current effect-eliminated bio-inspired neural network was proposed to guide the AUV navigation considering the effect of currents \cite{Zhu_2021}. A current correcting component was incorporated with the bio-inspired neural network to generate the paths. Each neuron in the network, the velocity and direction of robots are corrected for eliminating the current effect. Thus, the generated UUV path is robust and efficient.

The real-world ocean environment is complex and unknown. The onboard robot sensors were used for robot navigation with a limited detection range. The ultrasonic sensor was used to interpret the sonar data and update the map based on the dempster’s inference rule \cite{Zhu_2014unknown}. A potential field bio-inspired neural network (PBNN) was proposed to generate a safe path in underwater environments \cite{Cao_2018pbnn,Cao_2021pbnn}. The planned path keeps a safe distance to the obstacles, which could avoid the collisions for the underwater robot navigation.

Multi-AUV systems cooperation has received lots of interest due to the fact that groups of AUVs can work more efficiently and effectively compared with a single AUV. The main task of AUVs cooperation is to assign several targets to a team of AUVs and avoid obstacles autonomously in underwater environments.
Due to the similarity of multi-tasks assignment and self-organizing map (SOM) neural networks, many researchers have been applied the SOM approach to solve task assignment problems of multi-robotic systems \cite{Anmin_Zhu_2006,Zhu_2010tasks,Yi_2017SOM} and multi-AUV systems \cite{Daqi_Zhu_2013,Huan_Huang_2012}. However, the SOM-based methods require an ideal 2-D work environment without obstacles. An integrated biologically inspired SOM (BISOM) method was proposed to deal with collision-free and multi-AUV task assignment problems \cite{Zhu_2018BISOM}. After integrated the bio-inspired neural network method, the AUV is able to avoid obstacles and speed jumps. The ocean currents could influence the AUV navigation in the underwater environment. A velocity synthesis algorithm was integrated with the BISOM approach for optimizing the individual robot path in a dynamic environment considering the ocean current \cite{Cao_2015VS}.

The BISOM method is able to generate the shortest path for the multi-robotic systems in most situations. However, the update rule of the BISOM method ignores the effect of obstacles. Therefore, although the winner AUV is the shortest distance from the target, the obstacles could increase the movement of the winner robot. A novel biologically inspired map algorithm was proposed by Zhu \textit{et al.}\cite{Zhu_2021BINNmap} for changing the update rule.  The winner rule is not the shortest distance between target and AUV, whereas the winner rule becomes the maximum neural dynamic value in the neural activity values.

\subsubsection{Target Search}
The fundamental problem of target search for multi-AUV search systems is how to control all the vehicles to search to their target along the optimized paths cooperatively. The initial work on search was carried out by simplifying the search problem as an area coverage problem. As same as in cleaning robot application, the landscape of neural activity can guide the robot to search every unknown areas until the target was searched \cite{Rui_2014TR}. However, the coverage algorithm is not an efficient search algorithm as the robot power is wasted by unnecessary visiting positions. 
\begin{figure}[htb]
  \centering
  \includegraphics[width=0.95\textwidth]{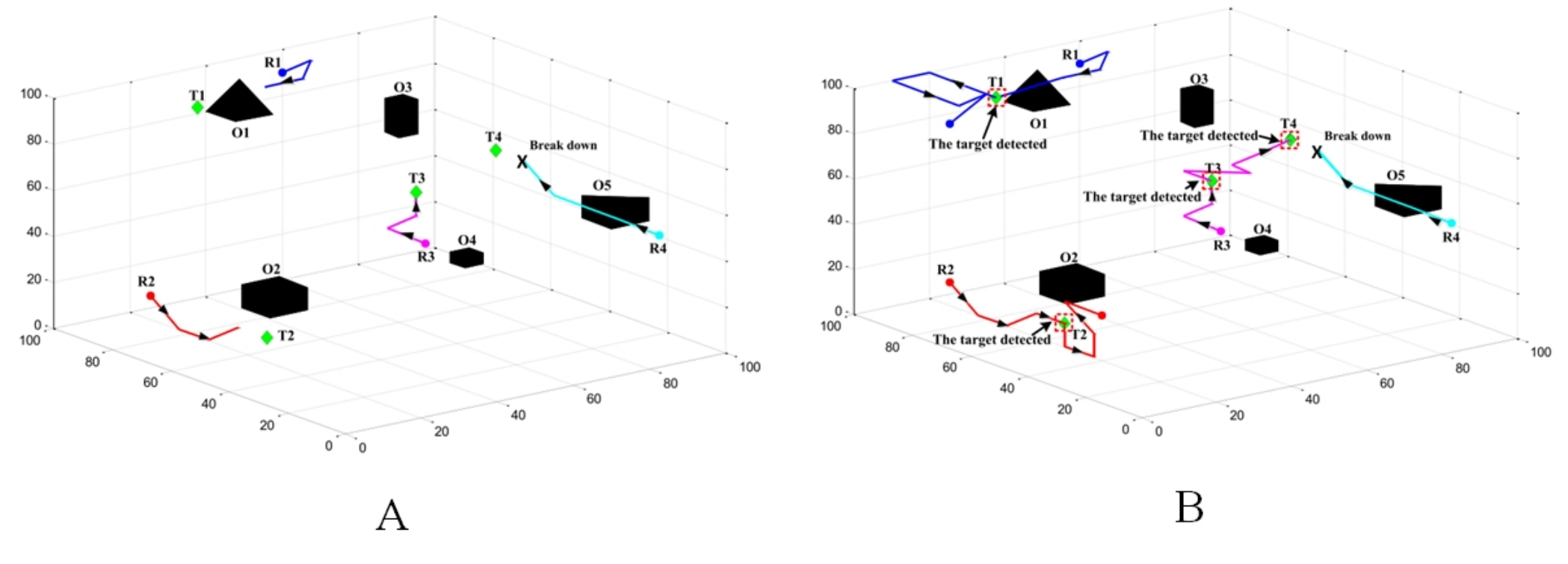}\hspace{1mm}%
  \caption{Examples of target search tasks. A: the AUV \textup{R4} breaks down; B: final trajectories of the search process \cite{Cao_2016TR}.}
  \label{fig:targetsearch}
\end{figure}
In order to improve the efficiency of the search algorithm, a sonar system was applied to extract the information of the environment to build the map and localize the target location \cite{Cao_2016TR}. \autoref{fig:targetsearch} shows that the proposed algorithm not only enabled the multi-AUV team to achieve search but also ensures a successful search if one or several AUVs fail. However, factors in real environments, such as ocean currents, were excluded in this simulation and there might be a waste of search capacity because of the overlapping search spaces. Same as the navigation application, with the consideration of ocean current, an integrated method based on the neurodynamics model and velocity synthesis algorithm was proposed for the cooperative search of the multi-AUV system \cite{Cao_2015ocean}.

\subsubsection{Hunting}
\begin{figure}[htb]
  \centering
  \includegraphics[width=0.65\textwidth]{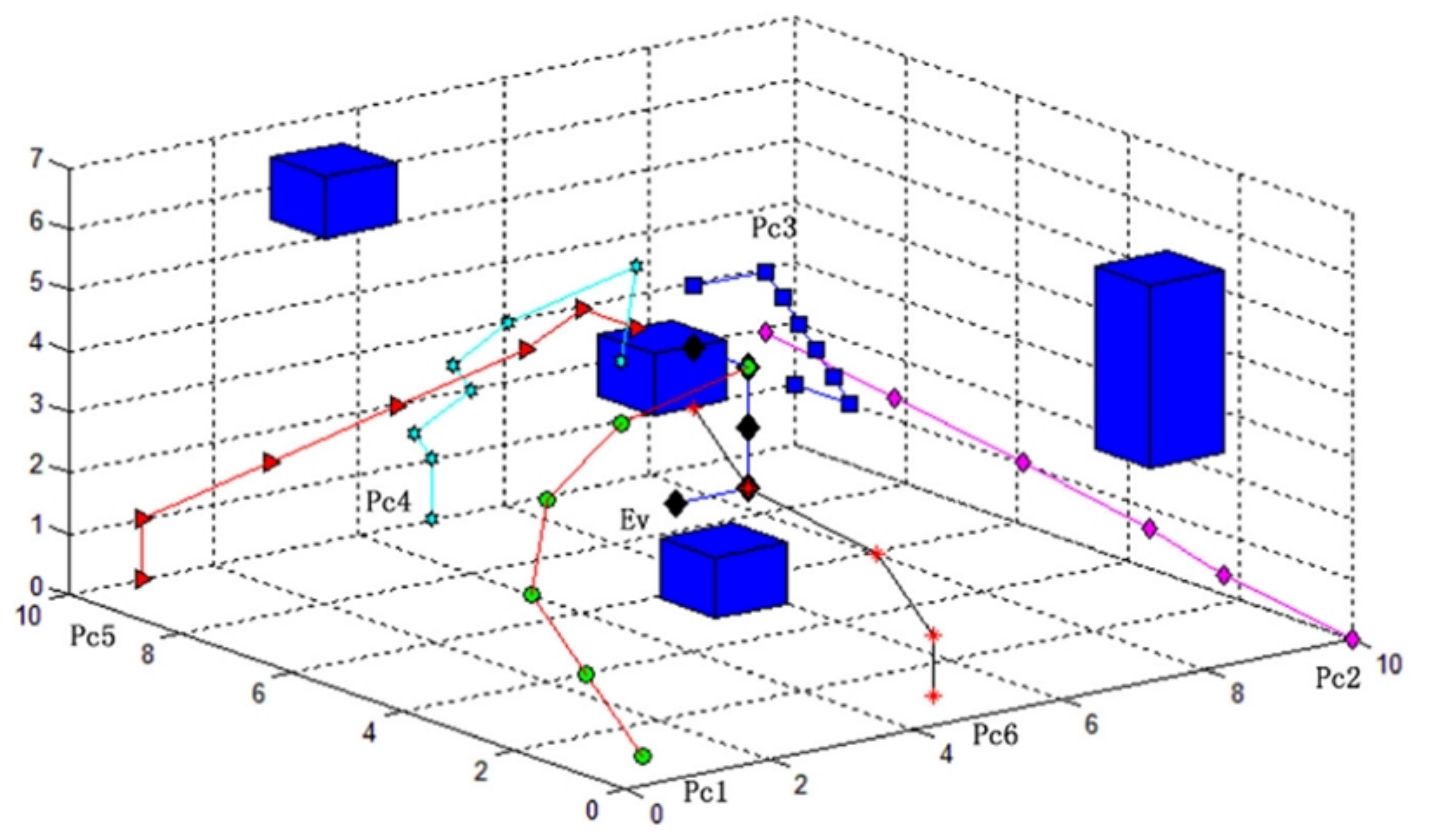}
  \caption{The hunting process with one target and six AUVs. $E_v$: the target; $P_{c}1$-$P_{c}6$: the AUVs \cite{Zhu_2015huntingauv}.}
  \label{fig:huntingAUV}
\end{figure}

Based on the previous study of neurodynamics model hunting for mobile robots in 2-D environments, a 3-D underwater environment hunting algorithm was proposed \cite{Huang_2015huntingauv,Zhu_2015huntingauv}. Compared with Ni and Yang's hunting algorithm \cite{Jianjun_Ni_2011hunting}, the catching stage was very different in applying underwater robots. The final hunting state can be divided into four situations. \autoref{fig:huntingAUV} shows one of the hunt situations that four AUVs surrounded the target. 

The path conflict situation happened when multiple AUVs chose the same position to be the next movement. A collision-free rule was established that location information is recorded between each AUV and selects the next step grid in each vehicle in anticipation before the movement \cite{Cao_2015huntingauv}. If any other AUV has occupied the grid, then choose another grid to move.

\section{Control}

Robot control is ongoing research that tracks much attention. The control in robotics is to develop controllers that drive robot kinematics or dynamics to reach desired states. Intelligent control of the robot is to develop a controller by taking advantage of vital characteristics of human intelligence, such as fuzzy logic, neural network, etc. bio-inspired intelligent control mechanism is based on biological systems. Using this biologically inspired system is targeting to improve the control performance by the implementation of natural biological systems in the control design.
\subsection{Tracking Control}
The tracking control of robots or motors has been studied for many years. Sliding mode control is robust to variable changes, however this method suffers chattering issues, which is a critical factor that needs to be considered when designing the control strategy. The linearization control method is easy to implement, however, it suffers from a large velocity jump when a large tracking error occurs at the initial stage. Backstepping control is easy to design, however, when a large tracking error occurs, this method becomes impractical as the speed jump will result in a large velocity surge, which can damage the hardware of the system. Neural network and fuzzy logic control are capable of resolving the large velocity jump at the initial stage, however, both neural network and fuzzy logic control are hard to practice. The neural network-based control methods require online learning, which is expensive and computationally complicate, the fuzzy logic control requires human experience to make the robot perform well, both of these control methods are rather expensive to practice.

\begin{figure}[h]
  \centering
  \includegraphics[width=1\textwidth]{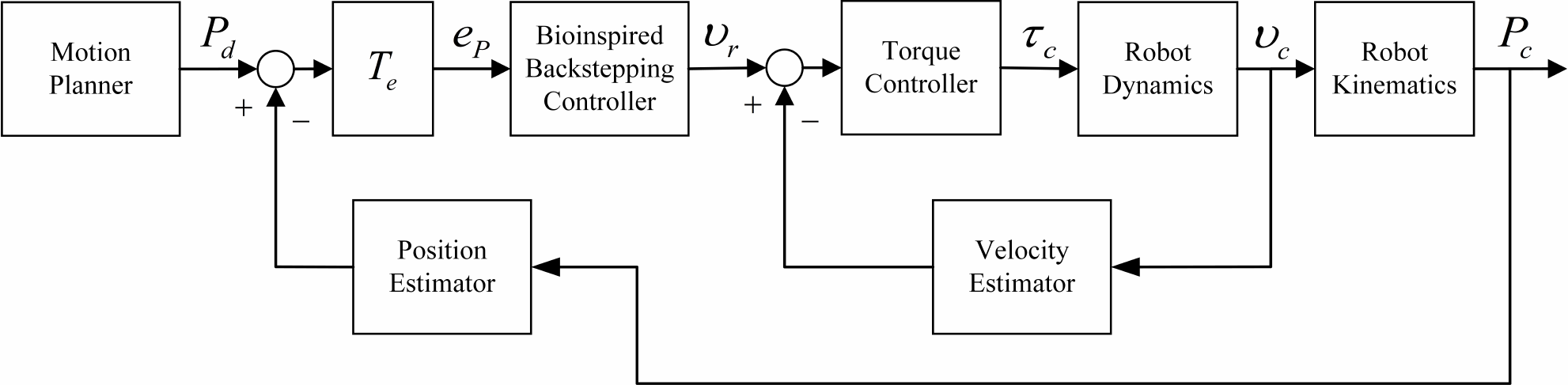}
  \caption{The block diagram of the bio-inspired tracking control for robots}
  \label{fig2}
\end{figure}

The bio-inspired backstepping control, which is based on the backstepping technique, aims to eliminate the speed jump in conventional design when a large initial tracking error occurs. The general control design for the unmanned robot with the implementation of bio-inspired neural dynamics can be described in \autoref{fig2}. The motion planner plans the desired posture $P_d$, then the desired trajectory along with the feedback of the current posture $P_c$ propagates through a transformation matrix to convert the tracking error from the inertial frame into body fixed frame. Then, the path tracker, which contains the bio-inspired backstepping controller uses the tracking error and desired velocity to generate a velocity command, which then along with the observed velocity $\upsilon_c$ propagate through torque controller to generate torque command, which drives the robot to generate a velocity and reach its desired posture by propagating the velocity that is generated from robot dynamics to robot kinematics. 

The applications of bio-inspired backstepping control are mainly divided into three different platforms: mobile robots, surface robots, and underwater robots. Therefore, this section illustrates the efficiency, effectiveness, and applications of the bio-inspired backstepping control into these three different platforms.
\subsubsection{Mobile Robots}
Real-time tracking control of a mobile robot is a challenging issue in mobile robotics. The main purpose of the tracking control is to eliminate or reduce the effects of errors. However, the disturbance, noise, and sensor errors will interfere with the output of the robotic system and produce errors. Many control algorithms of the mobile robot have been studied for precisely tracking a desired trajectory. The conventional backstepping control for mobile robots suffers from velocity jump issues, this problem is embedded in the design of the controller. The linear velocity error term causes the velocity jump if the initial tracking error does not equal zero. As seen this problem, the bio-inspired neural dynamics was brought into the design of the backstepping control.
For a nonholonomic mobile robot operates in a 2-D Cartesian work space, the main control variables for its kinematic model are the linear velocity and angular velocity. Focusing on the design of solving the velocity jump issue, the bio-inspired backstepping kinematic control for a mobile robot is defined as
\begin{equation}
\begin{array}{l}
{\upsilon _c} = {\upsilon _s} + {\upsilon _d}\cos {e_\theta }\label{37}
\end{array}
\end{equation}
\begin{equation}
\begin{array}{l}
   {\omega _c} = {\omega _d} + {C_1}{\upsilon _d}{e_L} + {C_2}{\upsilon _d}\sin {e_\theta }\label{38}
\end{array}
\end{equation}
\begin{equation}
\begin{array}{l}
\frac{d \upsilon_{s}}{d t}=-A \upsilon_{s}+\left(B-\upsilon_{s}\right) [e_{D}]^+-\left(D+v_{s}\right) {[e_{D}]^-}
\label{eq:tracking}
\end{array}
\end{equation}
where $\upsilon _s$ is derived from neural dynamics equation regards to the error in driving direction for the mobile robot, $C_1$ and $C_2$ are the designed parameters, $\upsilon_d$ and $\omega_d$ are respectively the desired linear and angular velocity that are given at path planning stage, $\upsilon_c$ and $\omega_c$ are respectively the linear and angular velocity commands that generated from the controller, and $e_{D}$ and $e_{L}$ are respectively the tracking error in driving and lateral directions \cite{Yang_2012real}. Compared to conventional design, the bio-inspired backstepping control takes the advantage of the shunting model that provides bounded smooth output.  

The bio-inspired backstepping controller resolved the problem of sharp speed jumps at the initial stage \cite{Yang_2002tracking,Yang2001tracking,Guangfeng_Yuan2001tracking}. The total design of the proposed control and path planning method were able to provide both real-time collision-free path and provide smooth velocity tracking commands for a nonholonomic mobile robot. However, the generated angular velocity seemed to suffer from sharp changes, therefore, the validation of the proposed control strategy is needed. In addition, the simulation environment is assumed as a simple environment with no obstacles. Zheng \textit{et al.}\cite{Zheng_2019tracking} proposed an adaptive robust finite-time bio-inspired neurodynamics control with unmeasurable angular velocity and multiple time-varying bounded disturbances. The outputs were smooth and the sharp jumps of initial values were decreased.

In real-world applications, the model input of the mobile robot may have errors, therefore, to overcome the problem of this abrupt change in the generated velocities caused by the model input errors, a fuzzy neurodynamics-based tracking controller, which incorporated fuzzy control to generate smooth velocities, was proposed \cite{Yanrong_Hu2003tracking}. The proposed control considered the model input error that consequently have impacts on the tracking error, which was further reduced using fuzzy logic to incorporate with the bio-inspired backstepping control. In addition, the shunting model also incorporated with PID controller to modify the error term, this control strategy provided a smooth velocity curve and more importantly, avoided impulse acceleration and torque, which could potentially damage the mechanical system \cite{Hui_Di_ZhangPID}.

In order to improve the efficiency and effectiveness of the bio-inspired backstepping control, the parameters of the control were determined using a genetic algorithm \cite{Hao_Li2003optim}. Tuning control parameters with the genetic algorithm provided better results than the implementation of bio-inspired backstepping control alone. Although the parameters tuned with the genetic algorithm provided satisfactory results, many other optimization methods could be used to choose the parameters, a comparison study could be tested to demonstrate the efficiency of the genetic algorithm. A biologically inspired full-state tracking control technique was proposed to generate smooth velocity commands \cite{Yang_2004full}. The proposed control considered both position error and orientation error as the control input and used the shunting model to constrain its output to reach its goal of providing a smooth velocity curve. There are still some improvements can be made as the path itself is not smooth but has sharp turns before it tracks its desired trajectory in a straight line tracking simulation. In addition to the simulation studies, successful implementation on a real mobile robot system demonstrates the effectiveness of the bio-inspired backstepping controller \cite{Yang_2012real}.  The experiment results showed that the robot tracked both the straight path and the circular path, and simulation results provided smooth velocity curves.
\begin{figure} 
 \centering
\includegraphics[width=0.9\textwidth]{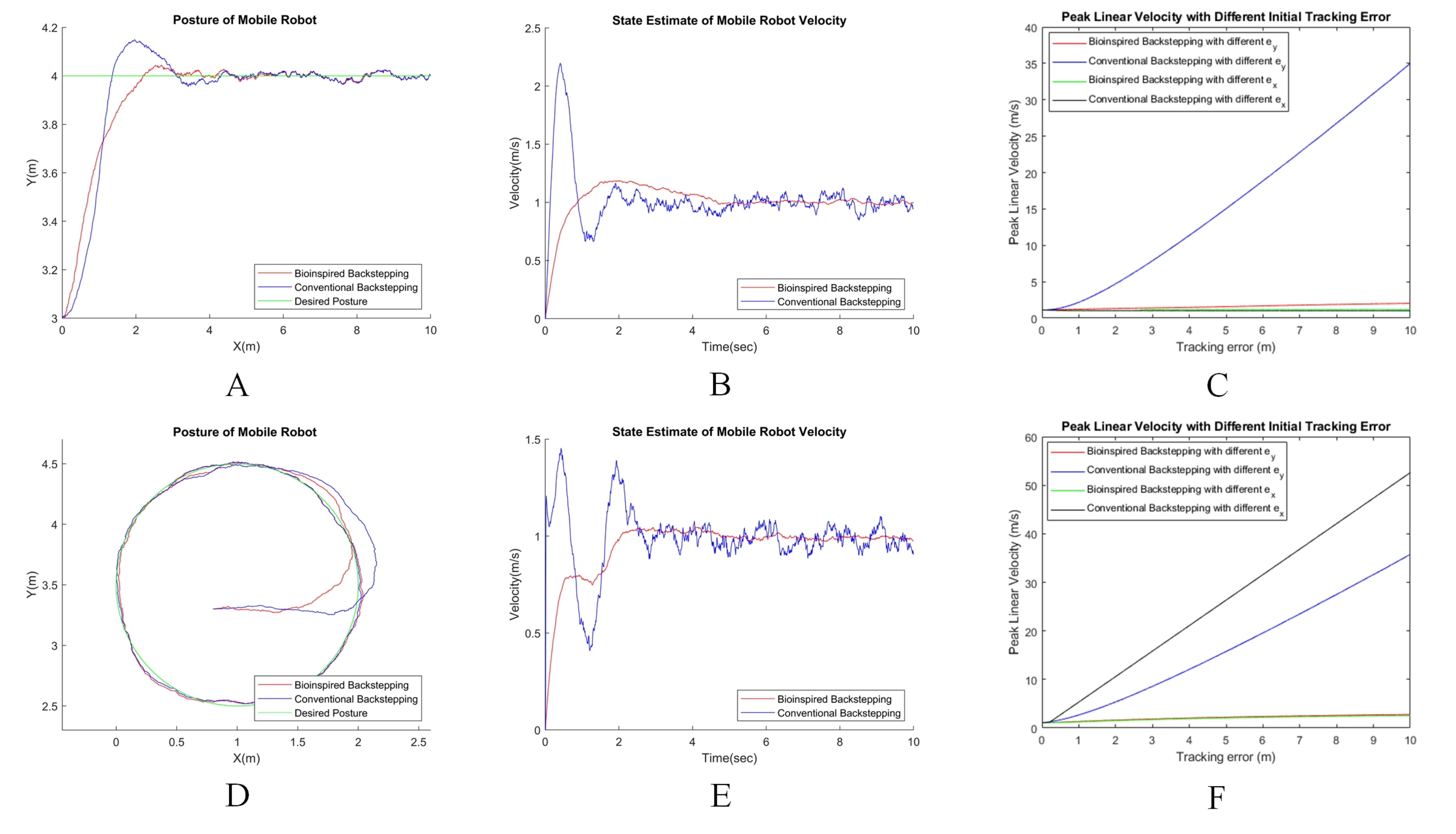}
  \setlength{\belowcaptionskip}{-0.4cm}
              \caption{The comparisons of the traditional backstepping control and bio-inspired backstepping control. A: tracking a straight line; B: linear velocity estimates of tracking a straight line; C: peak linear velocity comparison of tracking straight line; D: tracking a circular line; E: linear velocity estimates of tracking a circular line; F: peak linear velocity comparison of tracking a circular line \cite{Xu_2020ukf}.}
     \label{fig9}
\end{figure}

The mobile robot usually works in a complicated environment, which system and measurement noises can affect its tracking accurate. Therefore, an enhanced a bio-inspired backstepping control was proposed to generate the smooth, accurate velocity and torque command for mobile robots, respectively \cite{Xu_2020ukf}. The total control Incorporated bio-inspired backstepping controller with unscented Kalman and Kalman filters that were suitable in real-world applications. The proposed control considered noises in real-world applications, and the proposed control considered such noises effect and successfully eliminated it. However, the proposed control is considered a fixed noise, which is not true in real-world applications.

To illustrate the efficiency and effectiveness of the bio-inspired backstepping control for a mobile robot, \autoref{fig9} is chosen to show the superiority of the bio-inspired backstepping control over the conventional method. As seen from  \autoref{fig9}A   and \autoref{fig9}F, the larger the initial error occurs, the larger the initial velocity jump from the conventional method occurs, however, the bio-inspired backstepping control still makes the robot maintain a low initial velocity change. It is obvious that the bio-inspired backstepping control has practically solved a speed jump issue in backstepping control for a mobile robot, which is more practical in real-world applications.

\subsubsection{Surface Robots}
The tracking problem of the unmanned surface vehicle (USV) usually refers to the design of a tracking controller that forces robots to reach and follow a desired curve, where 2-D and three DOF (surge, sway and yaw) are considered \cite{8.22pan2015,8.22mohd2018}. 

\begin{figure}[htb]
  \centering
  \includegraphics[width=0.95\textwidth]{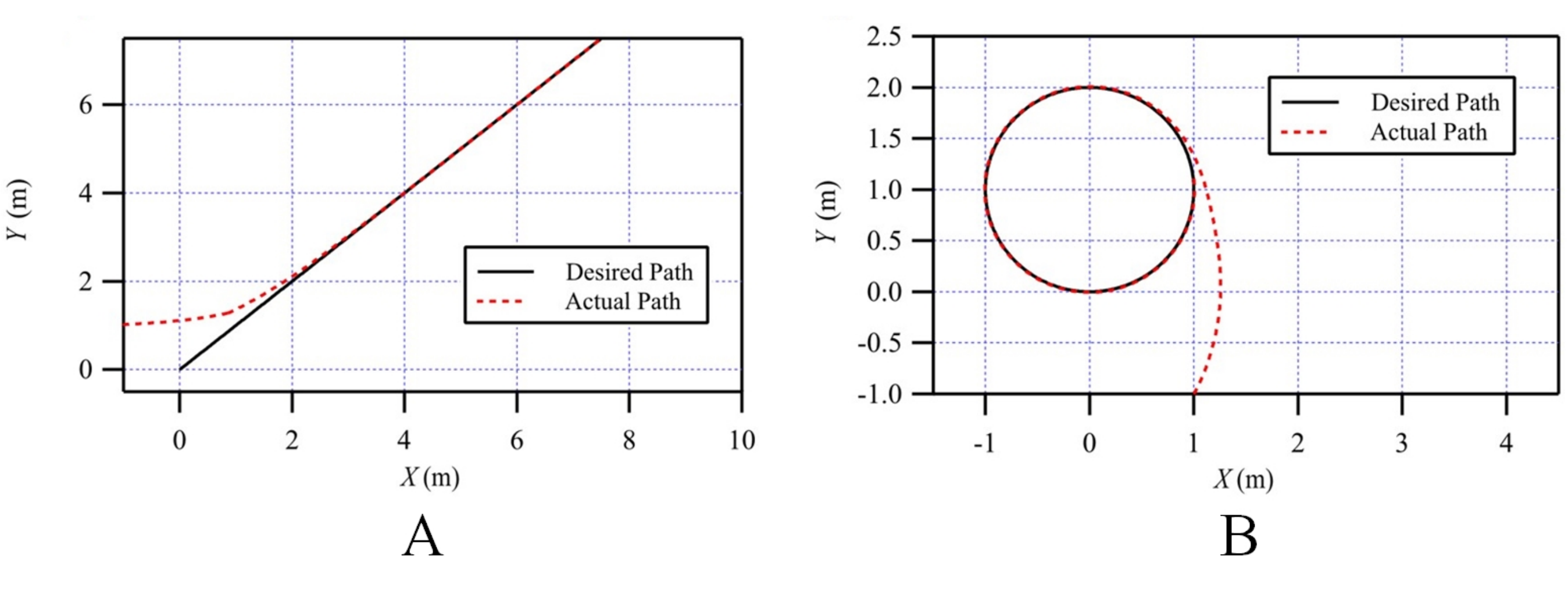}
  \caption{Tracking trajectory comparison of the bio-inspired method and conventional backstepping method for the underactuated surface vessel. A: line tracking; B: circle tracking \cite{8.22pan2015}.}
  \label{fig:trackUSV}
\end{figure}

The bio-inspired backstepping controller was used to USV for dealing with the velocity-jump problem \cite{Pan_2013ASV}. In the case that considering the impact of ocean current, a current ocean observer is fused with the control design to reduce the impact of ocean current in the tracking performance \cite{Pan_2015ocean}. The bio-inspired backstepping controller was integrated with a single-layer neural network for underactuated surface vessels in unknown and dynamics environments \cite{8.22pan2015}. The proposed tracking controller reduced the calculation process, therefore, the tracking controller avoided the complexity problem existed in conventional backstepping controllers. The stability of the tracking control system is guaranteed by a Lyapunov theory, and the tracking errors are proved to converge to a small neighborhood of the origin such that a satisfactory tracking result is presented in \autoref{fig:trackUSV}.

\subsubsection{Underwater Robots}
Bio-inspired neurodynamics models have been applied to the tracking control of underwater robots for many years \cite{8.22li2019}. The tracking control of the underwater robots is generally addressed by designing a control law that realizes asymptotically exact tracking of a reference trajectory based on the given underwater robots plant model \cite{8.22burdinsky2012}. However, different from common robots such as the land vehicle or the USV, the underwater robotics system contains more states, whose DOF can be extended to six. Among the six DOFs of the underwater robots, surge, sway, heave, roll, pitch, and yaw, roll and pitch can be neglected since these two DOFs barely have an influence on the underwater vehicle during practical navigation. Therefore, when establishing the trajectory tracking model to keep a controllable operation of the underwater robots, usually only four DOFs: surge, sway, heave, and yaw are involved. As same as the mobile robot, the speed jumps largely affect the robustness of the underwater robots path tracking. Due to the complex underwater work environment and limited electric power of underwater robots, the speed jumps as well as the driving saturation problem have to be considered. The bio-inspired backstepping controller was introduced in the control design to give the resolution respectively \cite{8.16karkoub2017}. Due to the characteristics of the shunting model, the outputs of the control are bounded in a limited range with a smooth variation \cite{Zhu_2013sliding}.

\begin{figure}[htb]
  \centering
  \includegraphics[width=0.95\textwidth]{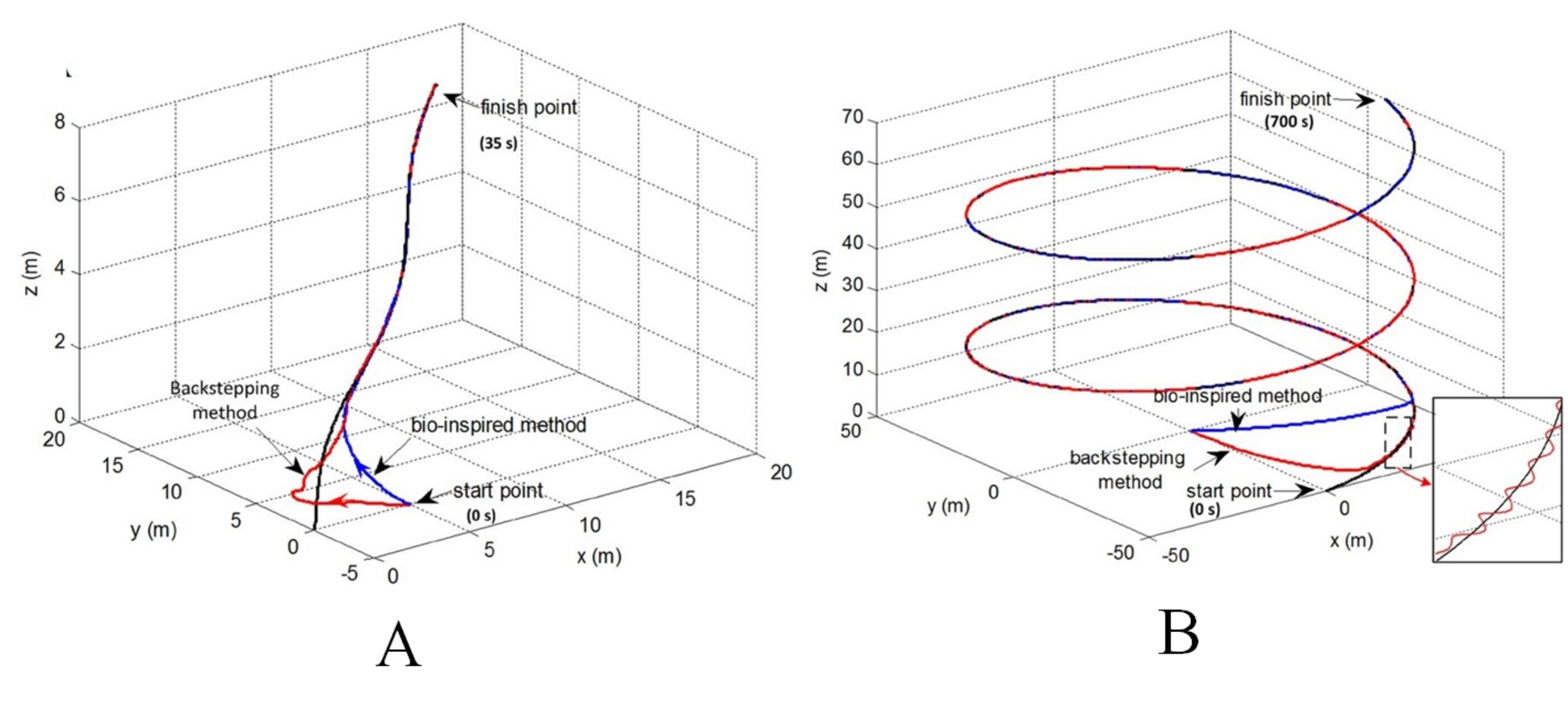}
  \caption{Tracking trajectory comparison of the bio-inspired model based method and conventional backstepping method for the underwater robots. A: curve tracking; B: helix tracking \cite{Sun_2014_7000m}.}
  \label{fig:tracking8.18}
\end{figure}

The bio-inspired backstepping controller has been applied on different underwater robots under various conditions by combining with a sliding mode control that controls the dynamic component of the vehicle, where an adaptive term is used in the sliding mode control to estimate the non-linear uncertainties part and the disturbance of the underwater vehicle dynamics \cite{Sun_2012sliding}. For example, the driving saturation problem of a 7000m manned submarine was resolved through this bio-inspired backstepping with the sliding mode control cascade control \cite{Sun_2014_7000m}. The control contains a kinematic controller that used bio-inspired backstepping control to eliminate the speed jump when the tracking error occurred at the initial state. Then, a sliding mode dynamic controller was proposed to reduce the lumped uncertainty in the dynamics of the underwater robots, thus realizing the robust trajectory tracking control without speed jumps for the underwater robots \autoref{fig:tracking8.18}. Jiang \textit{et al.}\cite{Jiang_2018ocean} accomplished the trajectory tracking of the autonomous underwater robots in marine environments with a similar bio-inspired backstepping controller and the adaptive integral sliding mode controller. In the sliding mode controller, the chattering problem was alleviated, which increased the practical feasibility of the vehicle. However, more studies are needed to compare to prove the effectiveness of the proposed control strategy, such as the tracking control based on the filtered backstepping method.

\subsection{Formation Control}
The bio-inspired neurodynamics trajectory tracking control for a single nonholonomic mobile robot can be extended to the formation control for multiple nonholonomic mobile robots, in which the follower can track its real-time leader by the proposed kinematic controller. This section introduces leader-follower formation control based on the bio-inspired neurodynamics tracking controller into three different robot platforms.

\subsubsection{Mobile Robots}
The leader-follower formation control based on the bio-inspired neurodynamics tracking controller was studied by Peng \textit{et al.} \cite{Peng_2013formation}. The asymptotic stability of the closed-loop system was guaranteed. The issue of impractical velocity jump arising from the use of the backstepping approach was handled by means of the bio-inspired neurodynamics model. However, the control design was based on the level of the kinematics model so that the performance of formation control relies highly on the low-level servo controller. The robustness property of the system is not analyzed, which is important when facing uncertainties or disturbances. Therefore, further improvements can be made based on these aforementioned points. A leader–follower formation control using a bioinspired neurodynamics-based approach was proposed by Yi \textit{et al.}\cite{Yi_2018formation} for resolving the impractical velocity jump problem of nonholonomic vehicles. Simulation results demonstrate the effectiveness of the proposed control law.

In order to further improve the tracking performance, a non-time based controller was also proposed \cite{Hu2002ge}. The path planner not only generated a desired path for the mobile robot, but also became part of the control to adjust the actual path and desired path. Along with the bio-inspired backstepping tracking control, the proposed method provided an overall better performance than a single backstepping control alone for multi-robotic systems.

\subsubsection{Surface Robots}
To fulfill the requirement of accomplishing complex tasks in the unpredictable marine environment, where the ocean currents and the marine organism may affect the efficiency of the vehicle operation, formation control on the system of multiple USVs has become a hot topic in recent decades \cite{8.26he2021}. Studies of combining the bio-inspired model with the marine vehicle formation control have been proposed and the model is often used to achieve the intelligent planning results of the multi-vehicle system \cite{8.26peng2021}.

Regarding the bio-inspired model application on the USV formation control, a novel adaptive formation control scheme based on bio-inspired neurodynamics for waterjet USVs with the input and output constraints was proposed \cite{Wang_2019formation}. However, the learning process of the adaptive neural network can reduce the real-time performance, which is the superiority of the bio-inspired neural network. In addition, the robustness property of the resulting closed-loop system is not analyzed when the undesired perturbation is injected into the system, which is considered a critical problem in practical engineering.

\begin{figure}[htb]
  \centering
  \includegraphics[width=0.8\textwidth]{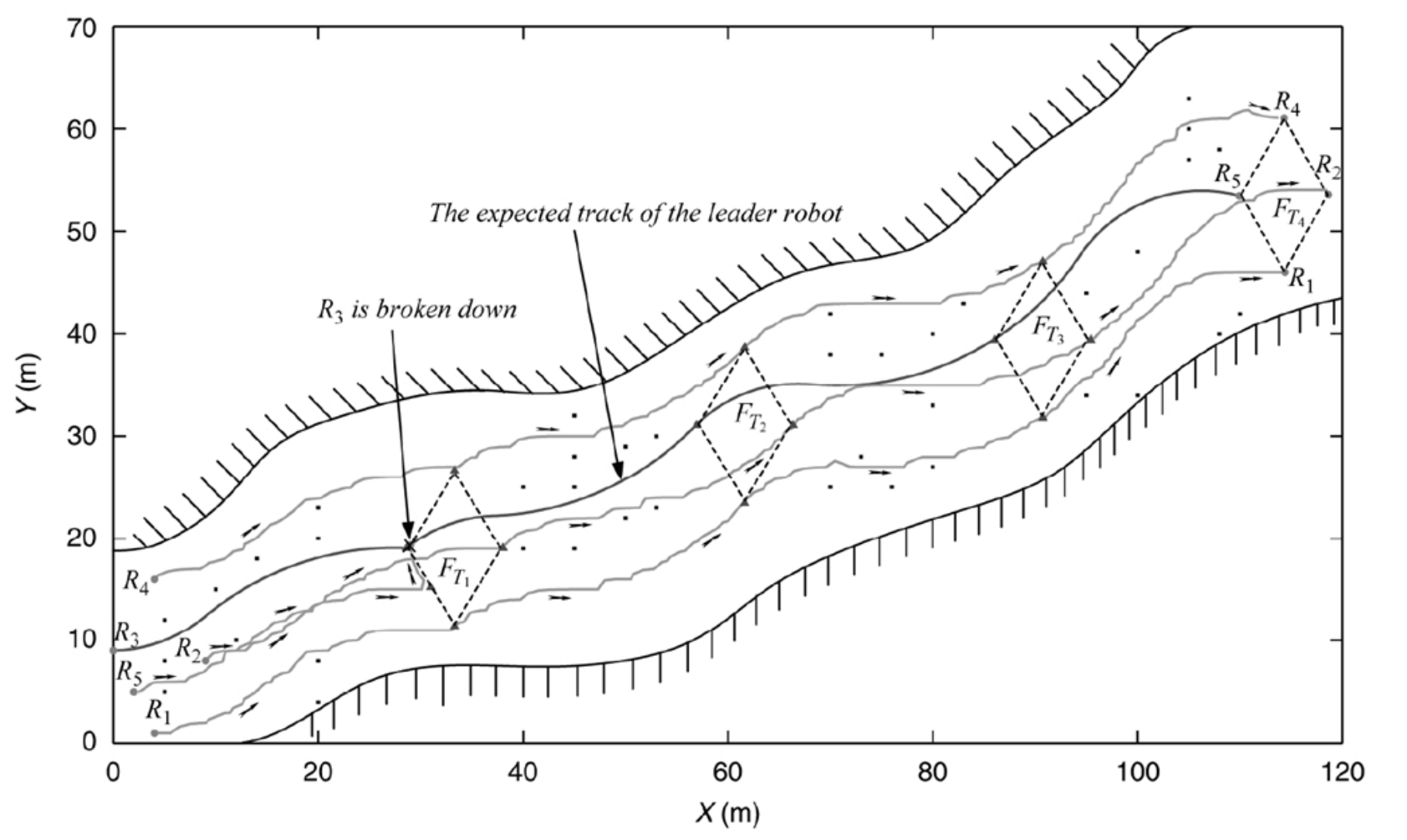}
  \caption{The simulation experiment in the case that the leader robot $R_3$ is broken down \cite{Ni_2015formation}.}
  \label{fig:leaderBroken}
\end{figure}

For multi-robotic system operates in large and unknown environments, Ni \textit{et al}. used a dynamic bio-inspired neural network for real-time formation control of multi-robotic systems in large and unknown environments \cite{Ni_2015formation}. The proposed approach considered many uncertain situations. \autoref{fig:leaderBroken} shows that the multi-robotic systems still finish the formation task, when the leader USV was broken. However, the mathematical analysis for the proposed algorithm is not provided, such as convergence analysis and robustness analysis. Comparison with traditional approaches is not provided, thus it is not sufficient to demonstrate the efficiency of the proposed method.

Intelligent formation control for a group of waterjet USVs considering formation tracking errors constraints was proposed \cite{Wang_2021formation}. To guarantee line-of-sight range and angle tracking errors constraints, a time-varying tan-type barrier Lyapunov function is employed. Besides, the bio-inspired neurodynamics was integrated to address the traditional differential explosion problem, i.e., avoiding the differential operation of the virtual control. However, the simulation example is much limited, thus the effectiveness and efficiency of the proposed control scheme are not sufficiently verified, i.e., the lack of the comparison with another type of control method.

\subsubsection{Underwater Robots}
For underwater robots, the definition of formation control is similar to the surface vehicle but with additional dimensions \cite{8.16yang2021}. Formation control of the multi-UUV system considers both 2-D and 3-D, where the former focuses on the lateral movement of the vehicle groups \cite{8.22hadi2021}.

A formation control on the multi-UUV system to realize the tracking of desired trajectory and obstacle avoidance in the 3-D underwater environment was proposed by Ding \textit{et al.} \cite{Ding_2014formation}. The bio-inspired neural network helps the leader UUV decide the transform of the formation when encountering obstacle fields to avoid the obstacles for all UUVs and meanwhile sustain on the desired trajectory. However, complex environmental disturbances such as multi-obstacles are not thoroughly considered in this paper.

\section{Challenges and future works}

Although there have been many studies of bio-inspired intelligence with applications to robotics and remarkable achievements have been accomplished, there are still several challenges that would be further investigated as future works.
\begin{itemize}
\item
Many existing approaches assumed the environment is static and without any uncertainties (e.g., some robot breakdown), disturbance (e.g., wind for surface robots, ocean/river current for underwater robots), and noise (system and measurement noise). However, it is a big challenge for collision-free robot navigation in complex changing environments with many moving robots/targets and subject to uncertainties, disturbance, and noise. 

\item
Communication issue has always been an essential research area in robotics. It is important to build a stable communication network in multi-robotic systems to ensure the updating of neural activity in the bio-inspired neural network. However, most studies on the cooperation of multi-robotic systems did not consider the communication issue, where the communication is normally noisy and with time delay. Many approaches did not consider the optimal performance with multi-objectives (e.g., short total distance, completion time, energy, smoothness of the robot paths). Communication and multi-objectives optimization could be a potential research direction in the future.

\item
Most conventional aerial robot navigation cannot act properly due to the limitations of communication and perception ability of sensors in complex environments. The complexity of the aerial robot makes the controllers are hard to design to achieve overall good performance. Though real-time collision-free navigation and control of mobile robots, surface robots, and underwater robots have been studied for many years, there is a lack of research for aerial robots based on bio-inspired neurodynamics models. The future research is to incorporate bio-inspired neurodynamics models with other useful algorithms for aerial robot navigation. 

\item
Most studies on the navigation and control of robots fail not to consider the teleoperation and telepresence issues. It is assumed that the robot works without human interactions. New approaches to telerobotic operations and human-robot interactions would be developed based on biologically inspired intelligence to outperform existing technologies. The future developed algorithms will not directly mimic any biological systems. The infusion of “human-like” and biological intelligence into robotic systems is the crux of future research.
\end{itemize}

\section{Conclusion}

Biologically inspired intelligence has been explored and studied for decades in the field of robotics. The researchers have been trying to replicate or transfer the biological intelligence to robotic systems for empowering the robots stability, adaptability, and cooperativeness. This paper provides a comprehensive survey of the research on bio-inspired neurodynamics models and their applications to path planning and control of autonomous robots. Among all bio-inspired neurodynamics models, shunting models, additive models, and gated dipole models were further elaborated. As for path planning, a bio-inspired neural network was elaborated for the dynamic collision-free path  generation for many robotic systems. There are several key points are worth to highlight about bio-inspired neurodynamics models to real-time collision-free path planning.
\begin{itemize}
\item The fundamental concept of the neurodynamics-based path planning approach is to develop a one-to-one correspondence neural network, which is called the bio-inspired neural network, to represent the work environment. The neural activity is a continuous signal with both upper and lower bounds.
\item The bio-inspired neural network is able to guide the robot to avoid the local minima points and the deadlock situations. The target globally influences the whole work space through neural activity propagation to all directions in the same manners.
\item The bio-inspired neural network is able to generate the path without explicitly searching over the free work space or the collision paths, without explicitly optimizing any global cost functions, without any prior knowledge of the dynamic environment, and without any learning procedures.
\item The bio-inspired neural network is able to perform properly in an arbitrarily dynamic environment, even with sudden environmental changes, such as suddenly adding or removing obstacles or targets. The obstacles have only local effects to push the robot to avoid collisions.
\end{itemize}
As for the bio-inspired robot control, several key points are worth to note:
\begin{itemize}
    \item The neural activity is bounded between the $[-D, B]$ region with different inputs, which is the fundamental concept of the bio-inspired backstepping control.
    \item The bio-inspired backstepping control provides a smooth velocity curve, which is crucial to ensure the control effectiveness and efficiency
    \item The speed jump in conventional backstepping control design is eliminated by replacing the tracking error term with shunting model, this modification allows a wider application of the bio-inspired backstepping control in robotics.
    \item The excellent feasibility of the bio-inspired backstepping control allows it compatible with many other control strategies to form new hybrid control strategies for robots working in various working environments.  
\end{itemize}
 The current challenges and future works are the development of original and innovative new intelligent navigation, cooperation, and communication strategies, algorithms and technologies with consideration of uncertainties, disturbance and noise issues, communication issues, and human-robot interaction issues for robots in changing complex situations.

  \bibliographystyle{IEEEtran} 
\balance
\bibliography{references}

\end{document}